\documentclass{article}
\usepackage[preprint,]{neurips_2026}

\usepackage[utf8]{inputenc} 
\usepackage[T1]{fontenc}    
\usepackage{hyperref}       
\usepackage{url}            
\usepackage{booktabs}       
\usepackage{amsfonts}       
\usepackage{nicefrac}       
\usepackage{microtype}      
\usepackage{xcolor}         

\usepackage{graphicx, subcaption}
\usepackage{colortbl}
\usepackage{tikz}
\usetikzlibrary{shapes.geometric}
\usetikzlibrary{arrows.meta}
\usetikzlibrary{positioning}
\usetikzlibrary{backgrounds, fit}
\usetikzlibrary{arrows.meta,calc}
\usepackage{amsmath}
\usepackage{amsthm}
\usepackage{amssymb}
\usepackage{xfrac}
\usepackage{stmaryrd}
\usepackage{multirow}
\usepackage{adjustbox}
\usepackage{float}
\usepackage{algorithm}
\usepackage{algorithmic}
\usepackage{enumitem}

\theoremstyle{plain}
\newtheorem{theorem}{Theorem}[section]
\newtheorem{proposition}[theorem]{Proposition}

\theoremstyle{definition}
\newtheorem{definition}[theorem]{Definition}

\theoremstyle{remark}
\newtheorem{remark}[theorem]{Remark}

\newcommand{\G}{\ensuremath{\mathcal{G}}}
\newcommand{\V}{\ensuremath{\mathcal{V}}}

\newcommand{\bA}{\ensuremath{\mathbf A}}

\newcommand{\RR}{\ensuremath{\mathbb R}}
\newcommand{\bT}{\ensuremath{\boldsymbol{\theta}}}

\newcommand{\bX}{\ensuremath{\boldsymbol{X}}}
\newcommand{\by}{\ensuremath{\boldsymbol{y}}}

\newcommand{\bg}{\ensuremath{\mathbf g}}

\newcommand{\bK}{\ensuremath{\mathbf K}}
\newcommand{\bZ}{\ensuremath{\mathbf Z}}
\newcommand{\bH}{\ensuremath{\mathbf H}}

\newcommand{\bM}{\ensuremath{\mathbf M}}

\newcommand{\bId}{\ensuremath{\mathbf{I}_d}}
\newcommand{\bIn}{\ensuremath{\mathbf{I}_n}}
\newcommand{\bE}{\ensuremath{\boldsymbol{\varepsilon}}}

\newcommand{\bS}{\ensuremath{\boldsymbol{\Sigma}}}

\definecolor{bluecb}{rgb}{0.45,0.31,0.49}
\definecolor{orangecb}{rgb}{0.88, 0.31, 0.24}
\definecolor{darkgray}{rgb}{0.15,0.15,0.15}
\definecolor{lightgray}{rgb}{0.94,0.94,0.94}
\definecolor{lightlightgray}{rgb}{0.97,0.97,0.97}
\definecolor{darkred}{rgb}{0.80,0.00,0.00}
\definecolor{darkgreen}{rgb}{0.00,0.70,0.00}
\definecolor{deepgreen}{rgb}{0.00,0.40,0.00}
\definecolor{darkblue}{rgb}{0.00,0.00,0.70}

\usepackage[normalem]{ulem}

\newcounter{marginNoteCounter}

\newcommand{\inlinetitle}[2]  {\smallskip\noindent\textbf{#1{#2}}}
\renewcommand{\paragraph}[1]  {\inlinetitle{#1}{}~}

\renewcommand*{\top}{{\mkern-1.5mu\mathsf{T}}}

\tikzstyle{startstop} = [rectangle, rounded corners, minimum width=3cm, minimum height=1cm, text centered, draw]
\tikzstyle{process} = [rectangle, minimum width=3cm, minimum height=1cm, text centered, draw]
\tikzstyle{arrow} = [thick,->,>=stealth]
\tikzstyle{arrowd} = [dashed,->,>=stealth]

\title{Cascaded Transfer: Learning Many Tasks \\under Budget Constraints}

\author{%
  Eloi Campagne$^{1,3}$\thanks{Corresponding authors: name.surname@ens-paris-saclay.fr} \And
  Yvenn Amara-Ouali$^{2,3}$ \And
  Yannig Goude$^{2,3}$ \And
  Mathilde Mougeot$^{1,4}$ \qquad Argyris Kalogeratos$^{1\ast}$ \\ [1.5em]
  $^{1}$ Centre Borelli, CNRS, ENS Paris-Saclay, Université Paris-Saclay, Gif-sur-Yvette, France \\
  $^{2}$ Laboratoire de Mathématiques d'Orsay, CNRS, Université Paris-Saclay, Orsay, France \\
  $^{3}$ EDF R\&D, Palaiseau, France \\
  $^{4}$ ENSIIE, Évry-Courcouronnes, France
}

\begin{document}

\maketitle

\begin{abstract}
In distributed applications, such as energy demand forecasting at the substation level or federated learning, a large number of related tasks must be learned by different models, while the exact task relationships are unknown.
We propose the novel \emph{Cascaded Transfer Learning} (CTL) paradigm in which model parameters cascade hierarchically through tasks organized as a rooted tree, respecting a global training budget. Starting from a source task, the tree specifies the order in which tasks are learned and refined, with the budget allocated along its branches.
We design cascade mechanisms based on spanning trees that connect all tasks by minimizing an objective combining pairwise task distances and the available training budget, which yield geometry-aware and depth-bounded transfer graphs. We theoretically characterize how transfer errors accumulate and attenuate along cascade paths: errors introduced at any upstream node are contracted by every downstream refinement, and balanced tree topologies bound this accumulation. Experiments on synthetic and real many-task settings, time-series forecasting and image classification, show that CTL enables more accurate and cost-effective adaptation across large task collections than alternative approaches, with the largest gains at the tightest budgets.
\end{abstract}

\section{Introduction}
\label{introduction}

Modern learning systems increasingly operate in settings where a large number of related tasks must be handled under various operational and computational constraints, e.g. data scarcity, computational power, fixed training budget. Large sensor networks, grids, or federated systems for personalized modeling are typical environments where each task is associated with only a small amount of localized data and independent training quickly becomes inefficient \citep{fallah2020personalized}.
Representative application domains include energy networks, climate analysis, transportation, and retail, in which thousands of related prediction problems arise in parallel \citep{laptev2018applied, he2019efficient, antoniadis2024hierarchical}. More specifically, there are large-scale spatiotemporal forecasting benchmarks, such as \texttt{SubseasonalClimateUSA} \citep{mouatadid2023subseasonalclimateusa} and \texttt{WeatherBench~2} \citep{rasp2024weatherbench}, which frame climate and weather prediction as collections of localized tasks evaluated under realistic computational constraints. Similar challenges appear in industrial hierarchical forecasting at scale in retail and energy systems \citep{sprangers2024hierarchical}, as well as in personalized federated learning, where each device constitutes a distinct task requiring local adaptation under heterogeneity, as exemplified by Ditto \citep{li2021ditto}.
In these many-tasks contexts, exploiting relationships across tasks is essential for building accurate models in frugal settings, as supported by both empirical and theoretical analyses in multi-task learning \citep{ciliberto2015convex}. To this end, scalable mechanisms are needed to exploit task relatedness across many heterogeneous tasks.

\begin{figure*}[t]
    \centering
    \includegraphics[width=\linewidth]{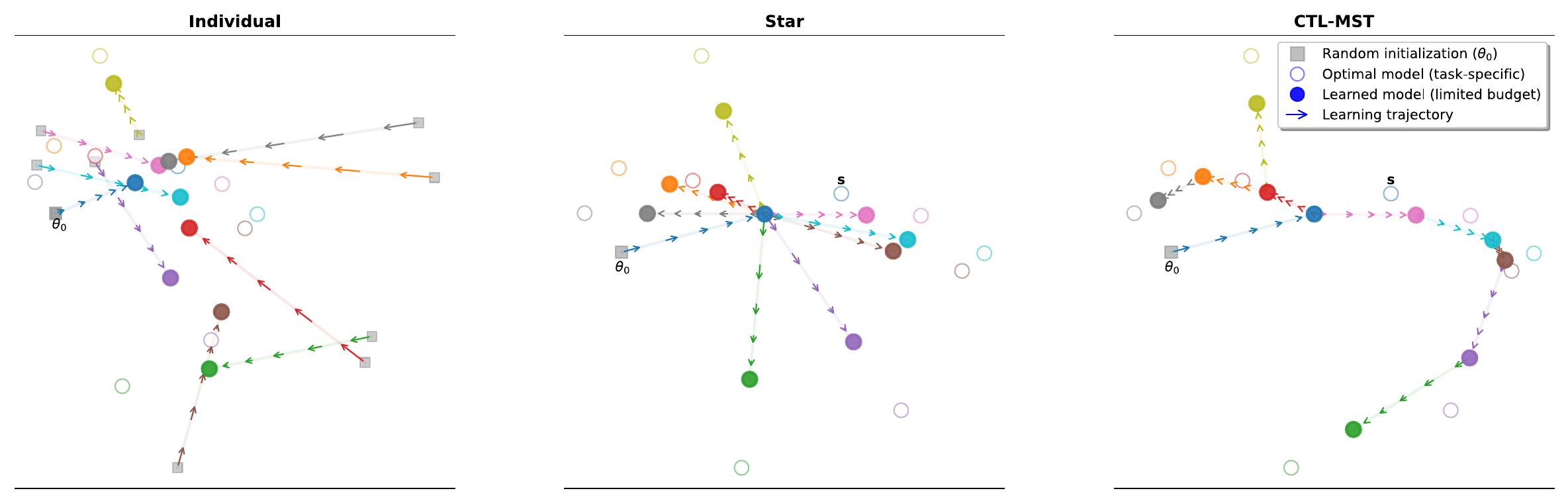}
    \caption{\textbf{Parameter-space intuition for CTL.}~ Arrows are shown along learning trajectories associated to different tasks (i.e. iterative parameter optimization). Each learning trajectory has a distinct color and stops at a learned model (solid nodes) that is short of its optimum (white-filled nodes) due to the limited available training budget.
    \emph{Left:} Independent training. Each task is optimized from its own initialization.
    \emph{Middle:} Star transfer. One source task is learned first and directly transferred to all other tasks.
    \emph{Right:} CTL where tasks are learned sequentially along a budget-aware spanning tree. Long transfers are decomposed into a series of short, locally corrected steps via intermediate tasks.}
    \label{fig:cascade-landscape}
\end{figure*}

\begin{table}[t]
\centering
\caption{Positioning CTL relative to MTL and TL in the frugal many-task learning regime.}
\label{tab:paradigms}
\resizebox{\columnwidth}{!}{
\begin{tabular}{llll}
\toprule
 & \textbf{Multi-Task Learning} & \textbf{Transfer Learning} & \textbf{Cascaded Transfer Learning} \\
\midrule
\textbf{Training} 
& Joint, iterative
& Independent per task
& Sequential cascade \\

\textbf{Information flow} 
& Shared parameters
& Source to each target
& Cascade over task graph \\

\textbf{Task updates} 
& Multiple (revisiting)
& Once (fine-tuning)
& Once (single-pass) \\

\textbf{Budget} 
& Joint training cost
& Independent per task
& Global shared allocation \\

\textbf{Interaction type}
& Implicit via shared parameters
& Direct source-to-target
& Explicit, distance-based task graph \\

\textbf{Error propagation} 
& Coupled via shared parameters
& Single-step bias
& Path-dependent cascade \\

\textbf{Scalability} 
& Limited (interaction overhead)
& High (parallel)
& High (sequential, parallel across branches) \\
\bottomrule
\end{tabular}}
\end{table}

\paragraph{Related work.} 
\emph{Multi-Task Learning} (MTL) has been developed to leverage shared information across tasks through joint training. Main approaches include feature-sharing architectures, low-rank parameterizations, and task clustering methods that explicitly model task relationships \citep{zhang2018overview,zhang2021survey,ruiz2024survey}. While effective, such methods require synchronized, globally coordinated optimization across tasks, leading to substantial memory or communication overhead. Moreover, MTL is highly sensitive to task relatedness \citep{standley2020tasks}, and inferring task relationships from data is often costly or assumption-heavy, motivating the use of \emph{a priori} structures.

\emph{Many-Task Learning} (MaTL) extends this setting to hundreds or thousands of tasks whose relatedness is only partially observable, often under strict resource constraints or at the cost of training very large models \citep{yu2024unleashing}. Scaling to this regime typically requires mechanisms that control parameter sharing and interference, such as task-specific routing \citep{strezoski2019many}, hierarchical architectures \citep{hashimoto2017joint, liu2021multi}, or transferability-based task selection \citep{tan2024transferability}. Furthermore, recent studies on large-scale multi-task pre-training underscore the dualities of both the potential and the practical complexity of this regime \citep{aribandi2022ext5}. Centralized joint optimization or training very large multi-task models quickly becomes computationally prohibitive and environmentally costly, a limitation also documented in federated optimization \citep{reddi2020adaptive}.

\emph{Transfer Learning} (TL) naturally enters the frame when thinking about sharing information from one task to another without global joint training. TL methods are commonly categorized as instance-based, feature-based, parameter-based, or relational \citep{pan2009survey,weiss2016survey,zhuang2020comprehensive}. In practice, TL is straightforward to deploy in modern neural settings and scales well to large numbers of downstream tasks, but it typically treats each source-target pair independently \citep{kornblith2019better}. As a result, large-scale TL systems often rely on simple star-shaped transfer schemes, where a single pretrained model is adapted separately to many tasks \citep{raffel2020exploring,kolesnikov2020big}, and classical TL provides little guidance on how to organize or coordinate transfers across a large collection of tasks. Taskonomy \citep{zamir2018taskonomy} addresses this gap by building a large-scale task-transfer graph over computer vision tasks and identifying beneficial source-target pairs, performing many-to-one transfer by concatenating representations from multiple sources into a single task-specific encoder. Task embedding methods such as Task2Vec \citep{achille2019task2vec} characterize individual task transferability via diagonal Fisher information embeddings and are used to predict pairwise transfer gain between a source and a single target. Task-ordering methods in continual learning \citep{li2025optimal,dohare2024loss} aim to mitigate forgetting and loss of plasticity by finding an optimal linear ordering of tasks for a shared global model trained sequentially. Meta-learning approaches such as MAML \citep{finn2017model,fallah2020personalized} optimize performance over a task distribution through a meta-training phase and require multiple inner-loop updates at test time.

\paragraph{Contributions and positioning.}
We focus on the frugal many-task setting, formalized as the MIMO (Multiple-Input Multiple-Output) regime \citep{yu2024unleashing}, in which each task is refined only once under a strict global budget. We propose \emph{Cascaded Transfer Learning} (CTL), a paradigm that lies at the middle ground between MTL and TL (Table~\ref{tab:paradigms}). Rather than adapting a source model directly to each target task, CTL propagates knowledge through a sequence of tasks organized as a spanning tree connecting all tasks (Figure~\ref{fig:cascade-landscape}). The key idea is that coordinated transfer structured as a budget-aware spanning tree, constructed by minimizing an objective over pairwise task distances and the available training budget, can offer gains over independent direct adaptations. 

To the best of our knowledge, CTL is the first graph-structured transfer learning framework in which each task is learned by its own dedicated model, as opposed to sharing a common representation. Compared to Taskonomy, CTL addresses collections of homogeneous tasks defined on distinct datasets, performs sequential one-to-one transfers in parameter space along a tree, and operates under a strict global refinement budget shared across all tasks. Unlike continual learning, each task in CTL has its own dedicated model refined exactly once, so forgetting and task interference are absent by construction. CTL is also complementary to meta-learning: it requires no meta-training phase and performs a single local refinement per task without iterative task revisiting.

Our contributions can be summarized as follows:
\begin{enumerate}[label=(\roman*),leftmargin=2em,itemsep=0em,topsep=0em]
    \item We provide a theoretical analysis of cascaded transfer over trees,
          characterizing how transfer errors accumulate and attenuate along
          cascade paths, and establishing sufficient conditions under which
          CTL provably improves over direct transfer.
    \item Building on this analysis, we present a scalable CTL algorithm that
          constructs a budget-aware spanning tree over tasks and allocates a
          global training budget along its branches. Experiments on synthetic
          and real many-task settings, time-series forecasting and image
          classification, show that CTL enables more accurate and cost-effective
          adaptation across large task collections than alternative approaches,
          with the largest gains at the tightest budgets.
\end{enumerate}

\section{Cascaded Transfer Learning}
\label{sec:CTL}

This section formalizes CTL, introduces the underlying task-dependency structure, and describes the algorithmic template that can be used in practice for deploying CTL. 

\subsection{Preliminaries}
\label{sec:preliminaries}

Let $\V$ be a set of tasks, with cardinality $|\V|$. For two tasks $u,v\in\V$, we write $u\!\to\! v$ to denote a transfer from task $u$ to task $v$. Let $n\in\mathbb{N}^\ast$ be the number of samples available and $d\in\mathbb{N}^\ast$ the feature dimension, such that for each task $v\in\V$, $\bX_v\in\mathbb{R}^{n\times d}$ denote the task-specific data matrix. Each task is associated with an unknown parameter vector $\bT_v^\star \in \mathbb{R}^d$ to be learned from data. Let $b_v \in \mathbb{N}^\ast$ denote the computational budget allocated to that task, i.e. a fixed amount of local optimization effort, measured in refinement steps. Similar notions of computational budget have been used in prior work on learning under constrained optimization effort \cite{wong2021leveraging}. 

Let $G_v^{b} : \mathbb{R}^d \to \mathbb{R}^d$ be a refinement operator corresponding to $b$ iterations of a gradient-based optimization algorithm applied to a parameter vector. Given a source task $u$ and a target task $v$, a transfer $u \to v$ consists of initializing the parameters of $v$ using information from $u$, followed by local refinement via $G_v^{b}$. This source--target adaptation forms the elementary building block of our framework, which extends such transfers to a graph-structured setting. In that regard, a central object is the \emph{rooted tree} $\mathcal{T}_s=(\mathcal{V},\mathcal{E},s)$, where $\mathcal{E}$ is the set of edges and $s$ the \emph{root}. A rooted tree is a \emph{directed acyclic graph} (DAG) where each node has at most one parent, denoted by $\mathrm{pa}(v)$. For convenience, we make the convention that $\mathrm{pa}(s)$ is a dummy node corresponding to a model with random parameters $\bT_\mathrm{init}$.

\subsection{The Cascaded Transfer Learning Paradigm}
\label{sec:CTL-def}
We now formalize CTL, starting from the rooted tree setting.

\begin{definition}\label{def:CTL}
\emph{Cascaded Transfer Learning} is a learning process in which tasks in a set $\mathcal{V}$ are learned sequentially following the structure of a rooted tree $\mathcal{T}_s\!=\!(\mathcal{V},\mathcal{E},s)$ that connects them. Starting from the root $s$, each subsequent task is initialized by its parent and is then locally refined.
\end{definition}

Figure~\ref{fig:cascade-landscape} illustrates the rationale of the CTL framework, and how, starting from a chosen seed, information is propagated from task to task along the cascade. Algorithm~\ref{alg:abstract_ctl} gives the pseudocode that involves three central design choices:
\begin{enumerate}[label=(\roman*),leftmargin=1.5em,itemsep=0em,topsep=0em]
    \item \emph{Seed selection.} The root task $s$ that initiates the cascade.
    \item \emph{Graph construction.} Building a rooted tree from a weighted task graph $\mathcal G$, whose edge weights encode pairwise task distances and determine the direction and locality of information flow.
    \item \emph{Budget allocation and refinement.} The assignment of refinement budgets
    $\{b_v\}_{v\in\V}$ and the execution of local optimization along the cascade.
\end{enumerate}

\begin{algorithm}[t]
    \small
    \caption{Cascaded Transfer Learning}
    \label{alg:abstract_ctl}
    \begin{algorithmic}[1]
    \STATE \textbf{Input:} set of tasks $\mathcal{V} = \{v_1, v_2, \dots\}$, total training budget $B$, undirected task graph $\G$
    \STATE \textbf{Output:} refined models $\{\tilde{\bT}_v\}_{v \in \mathcal{V}}$
    \vspace{1mm}\hrule\vspace{1mm}

    \STATE \textbf{Seed selection:} choose the root task $s \in \mathcal{V}$
    \STATE \textbf{Tree construction:} construct a rooted tree $\mathcal{T}_s = (\mathcal{V}, \mathcal{E}, s)$ from the graph $\G$
    \STATE \textbf{Budget allocation:} assign 
		budgets $\{b_v\}_{v \in \mathcal{V}}$, $\sum_{v\in\V} b_v = B$

    \STATE \textbf{Cascaded transfer:} 
		\STATE \quad initialize $\bT_{\mathrm{pa}(s)}$ with random $\bT_\mathrm{init}$
        \STATE \quad \textbf{for} each task $v \in \mathcal{T}_s$ in topological order \textbf{do}
        \STATE \qquad set $\bT_v^{(0)} = \tilde{\bT}_{\mathrm{pa}(v)}$
        \STATE \qquad \textbf{Refine:} $\tilde{\bT}_v = G_v^{b_v}\!\big(\bT_v^{(0)}\big)$
        \STATE \quad\textbf{endfor}
    \STATE \textbf{Return:} $\{\tilde{\bT}_v\}_{v \in \mathcal{V}}$
    \end{algorithmic}
\end{algorithm}

\paragraph{Cascade Construction Strategies.}%
Rather than treating transfers independently, CTL organizes them along a predefined graph structure. Consider three tasks $(S,I,T)$ with $S$ already learned. Direct transfers $S\!\to\!I$ and $S\!\to\!T$ may be suboptimal when learning $I$ brings the process closer to $T$, since the path $S\!\to\!I\!\to\!T$ decomposes a long transfer into shorter steps, and the refinement of $I$ produces a better initialization for $T$. Section~\ref{sec:ctl-theory} formalizes when such routing reduces transfer error. Rooted trees provide a simple realization of this principle: they are acyclic, connect all tasks, and support parallel execution per branch.

Tree construction depends on the input task graph $\mathcal G$ encoding pairwise task relationships. If an application-specific graph (e.g. spatial proximity, network topology) is available, it can be used directly; otherwise, task relationships must be inferred from data via distance proxies. Section~\ref{sec:construction} specifies which tree structures 
 satisfy these principles and describes practical construction algorithms.

\paragraph{Extensions.}%
CTL can be extended to cascade forests (multiple seeds over disjoint task subsets), general DAGs allowing multi-parent fusion. Moreover, another direction is the investigation of the interplay between a considered budget allocation strategy and the objective with which the spanning graph structure should be built. This issue is discussed in Section~\ref{sec:construction}. Besides, as we show theoretically in Section~\ref{sec:ctl-theory}, controlling the worst-case path length is important, and suitable structures such as the minimum-bottleneck spanning trees or others can be examined. The parameter-space analysis extends to multiclass classification under the Polyak-Łojasiewicz condition, and to adapter-based transfer (LoRA, prefix tuning) where the restricted loss landscape satisfies the contraction assumption locally; see Section~\ref{sec:limitations} for further discussion.

\subsection{Parameter-Space Analysis}
\label{sec:ctl-theory}
We justify CTL over rooted trees, showing: (i) cascades can improve over direct transfer by routing long-range transfers through shorter intermediate steps, (ii) only mild structural conditions are required, and (iii) MSTs naturally satisfy these conditions under locality.

\paragraph{Additional notations.}%
The Euclidean and Frobenius norms are denoted by $\lVert\cdot\rVert$ and $\lVert\cdot\rVert_F$. The positive definiteness of a symmetric matrix $\mathbf S$ is expressed as $\mathbf S\succ 0$. With $B \in\mathbb N^\ast$ we denote the total learning budget. Given a rooted tree $\mathcal{T}_s$ and a budget allocation $\{b_v\}_{v\in\V}$ satisfying $\sum_{v\in\V} b_v=B$, CTL refines tasks following the topological (root-to-leaf) order, with updates of the form  $\tilde\bT_v = G_v^{b_v}\bigl(\tilde\bT_{\mathrm{pa}(v)}\bigr)$.

For each task $v\in\V$, we associate a loss $\mathcal L_v:\RR^d\to\RR$ and a minimizer $\bT_v^\star \in \RR^d$. Let $\eta \in (0,1)$ be the learning step associated to the refinement operator $G_v$. We assume a contraction property in the parameter space: for all $v\in\V$, there exists a contraction rate $\rho_v\in(0,1)$ such that, for all $(\bT, b) \in \RR^d \times \mathbb{N}^\ast$,
\begin{equation*}
    \left\lVert G_v^b(\bT)-\bT_v^\star\right\rVert\le \rho_v^b\left\lVert\bT-\bT_v^\star\right\rVert.
\end{equation*} 
This condition ensures that each local refinement step reduces estimation error at a geometric rate. It holds, for instance, for gradient descent on strongly convex and smooth objectives, and serves as a sufficient condition for controlling error accumulation along a cascade.
Task discrepancy is measured by $d(u,v) = \bigl\lVert\bT_u^\star - \bT_v^\star\bigr\rVert$, the parameter-space distance between task optima. Although not directly observable, it quantifies the bias induced when transferring from $u$ to $v$: under the contraction assumption, initializing task $v$ from $\tilde\bT_u$ yields estimation error proportional to $d(u,v)$ up to a contraction factor.

\begin{proposition}[Cascaded transfer over a path]\label{prop:propagation-param}
For any task $v \neq v_0$, and a path $(v_0\!\to\!v_1\!\to\!\cdots\!\to\!v_m\!=\!v$), involving $m >1$ intermediate nodes, it holds:%
\begin{equation*}
    \bigl\lVert\tilde\bT_v-\bT_v^\star\bigr\rVert\le P_{1:m}\bigl\lVert\tilde\bT_{v_0}-\bT_{v_0}^\star\bigr\rVert+\sum_{i=1}^m P_{i:m} d(v_{i-1},v_i),
\end{equation*}%
where $\displaystyle P_{i:m} =\prod_{j=i}^m\rho_{v_j}^{b_{v_j}}$ is a multiplicative attenuation factor.
\end{proposition}
The contribution of each edge is progressively attenuated by downstream refinement. CTL improves over direct transfer when: (i) edges connect nearby tasks, (ii) the structure is acyclic and ordered, and (iii) budgets prevent error accumulation along deep paths. MSTs enforce locality, are acyclic by construction, and are noise-stable, making them a robust default.

\paragraph{CTL vs.\ TL.} The following result formalizes when routing transfer through intermediate tasks reduces the error induced by task mismatch under a fixed budget.

\begin{theorem}
\label{thm:ctl-vs-star}
Suppose a path $(s\!\to\!v_1\!\to\!\cdots\!\to\!v_m\!=\!v)$ starting from an exact task, $\tilde\bT_s=\bT_s^\star$, and having uniform refinement budget $b$ at each other node. Let $\delta_i=d(v_{i-1},v_i)$\,, $\delta_{\max} = \max_{1\le i \le m} \delta_i$\,, and $\rho_{\max} = \max_{1\le i \le m} \rho_{v_i}$\,.

The estimation errors of the cascaded transfer along the path and that of direct transfer $s\rightarrow v$ are:%
\begin{align*}
   \left\lVert\tilde\bT_v^{\mathrm{_\textsc{CTL}}}-\bT_v^\star\right\rVert \le \sum_{i=1}^m P_{i:m}\,\delta_i\,, \quad
    \left\lVert\tilde\bT_v^{\mathrm{_\textsc{TL}}}-\bT_v^\star\right\rVert
\le \rho_v^{b}\, d(s,v)\,.
\end{align*}
CTL has a tighter upper bound than TL when: 
\begin{align*}
    \delta_{\max} (1-\rho_{\max}^{mb}) < d(s,v)(1-\rho_{\max}^{b})\,.
\end{align*}%
\end{theorem}

Theorem~\ref{thm:ctl-vs-star} formalizes why cascades reduce transfer bias. Under the star scheme the error at the target is proportional to the full distance $d(s,v)$ contracted by a single refinement. Under CTL, the path is subdivided into $m$ shorter edges, and the contribution of edge $i$ is $P_{i:m}\,\delta_i$ rather than $\delta_i$ alone, where $P_{i:m}=\prod_{j=i}^m\rho_{v_j}^{b_{v_j}}$ is the \emph{downstream attenuation}: every downstream node refines away the inherited mismatch, so errors introduced early are contracted by the full sequence of subsequent refinements. Errors near the leaf, which have fewer downstream contractions, are also the shortest edges under a locality-aware construction.

The sufficient condition $\delta_{\max}(1-\rho_{\max}^{mb}) < d(s,v)(1-\rho_{\max}^{b})$ is satisfied when the longest cascade edge, scaled by the ratio of cumulative to single-step attenuation, is shorter than the direct distance. Since MSTs enforce $\delta_{\max} \ll d(s,v)$ for smooth task geometries, the condition holds broadly, explaining why locality-preserving structures are effective CTL defaults.

\begin{remark}\label{rem:scope}
Theorem~\ref{thm:ctl-vs-star} compares relaxed upper bounds: both the CTL and TL bounds replace individual contraction rates $\rho_{v_i}$ by the path-wise maximum $\rho_{\max}$, so the condition identifies when CTL's worst-case bound beats TL's worst-case bound. If the target task has $\rho_v \ll \rho_{\max}$, TL's exact bound $\rho_v^b\,d(s,v)$ may be tighter even when the condition holds. The theorem also assumes $\tilde\bT_s=\bT_s^\star$ to isolate transfer bias; in practice the root is refined by $b_s$ steps from a random initialization, adding a residual $\rho_s^{b_s}\|\bT_{\mathrm{init}}-\bT_s^\star\|$ that enters any descendant's bound multiplied by $P_{1:m}$, but this term is doubly contracted along the path and becomes negligible for moderate cascade depth.
\end{remark}

The parameter-space analysis extends directly to the linear feature-space setting (gradient descent on quadratic objectives induces a contraction toward the task-specific OLS solution) and to noisy observations. Full statements and proofs are given in Appendix~\ref{app:tree_proofs}.

\section{Cascade Construction Strategies}
\label{sec:construction}
Guided by the analysis of Section~\ref{sec:ctl-theory}, we now discuss how cascade trees can be constructed. Three design principles emerge: (i) prefer short edges to enforce transfer locality, (ii) avoid deep paths to bound error accumulation, and (iii) cover all tasks with a connected structure. We first compare trees and chains in light of the preceding analysis, then describe the default MST-based construction, and finally we present a refined budget-aware extension.

\subsection{Trees vs.\ Hamiltonian chains}
Proposition~\ref{prop:propagation-param} implies that trees have a structural advantage that grows with $T$. In a Hamiltonian chain of $T$ tasks, which is a degenerate one-branch tree with depth $T-1$, the attenuation of edge $k$ is $P_{k:T} = \rho_{\max}^{(B/T)(T-k)} \to 1$ as $k \to T$. This means that late-chain edges receive vanishing downstream correction. Contrary, a balanced tree of depth $\mathcal{O}(\log T)$ attenuates the same edge by $\rho_{\max}^{(B/T)\log T}$, which is an exponential improvement. Under the subtree-weighted log-edge (SLE) allocation $b_v \propto |s_v|\log(1+d(\mathrm{pa}(v),v))$, derived from KKT conditions on the cascade error surrogate and used as the default throughout, the late-chain attenuation becomes $(1+\delta_k)^{-\kappa}$ with $\kappa = -\log\rho_{\max}$, which no longer collapses to $1$, but the residual error from a large inter-cluster edge still scales as $\delta_k^{1-\kappa}$, and thus increases in $\delta_k$ when $\kappa < 1$. In a balanced tree under SLE, the budget at depth $\ell$ scales as $(T/2^\ell)\log(1+\delta_\ell)$: edges covering $\mathcal{O}(T/2^\ell)$ descendants receive proportionally more budget, whereas SLE applied to a chain concentrates budget at root-proximate nodes, mismatching the chain optimum (Appendix~\ref{app:budget_derivation}). For a $K$-cluster task geometry, a tree can reach each cluster via an independent branch, exposing no task to more than one inter-cluster hop. Trees dominate chains most strongly in such settings, 
while the gap narrows when tasks lie close together; see Table~\ref{tab:simplified-opt}. In this latter case, the exact order of tasks in the chain becomes also less important, as different Hamiltonian chains can have similar effects in terms of task learning.

\subsection{Default Tree Construction}%
CTL requires as input a pairwise task distance matrix and a rooting strategy. Task geometry is not directly observable; we use the gradient-based distance (Euclidean distance between $\ell_2$-normalized gradient vectors $\bg_v = \bX_v^\top \by_v$ at initialization), which consistently yields the best downstream performance among the proxies evaluated (Appendix~\ref{app:diagnostics}). An MST on this distance matrix connects all tasks while minimizing total edge weight, enforcing transfer locality by construction. The undirected MST structure is root-invariant \citep{kleinberg2006algorithm}; rooting determines the cascade ordering and affects estimation error through path depth. We use the medoid (the task minimizing total distance to all others) as root: it is computable from the distance matrix before the tree is built, ensuring a uniform, topology-agnostic selection across all compared methods. Among three strategies tested on synthetic tasks, a depth-central root yields lower MSE, but requires the tree topology first; the medoid offers a robust data-driven compromise (Appendix~\ref{app:diagnostics}).

\paragraph{Computational Complexity.}%
MST construction from a dense pairwise distance matrix costs $\mathcal{O}(|\mathcal{V}|^2\log|\mathcal{V}|)$, incurred once before the cascade. Training then scales linearly with the global budget $B$: each task is refined exactly once from its parent, so the dominant deployment cost is budgeted local updates, not joint or repeated optimization.

\subsection{Budget-Aware Construction}
\label{sec:mstc}

\begin{figure}[t]
    \centering
\begin{minipage}[c]{0.43\linewidth}
    \begin{tikzpicture}[
        task/.style={circle, draw=black, fill=white, minimum size=4mm, inner sep=1pt, font=\small},
        root/.style={circle, draw=black, fill=gray!30, minimum size=4mm, thick, font=\small, inner sep=1pt},
        future/.style={circle, draw=gray!130, fill=gray!10, minimum size=4mm, inner sep=1pt},
        edge/.style={-{Stealth[length=2mm]}, thick},
        candidate/.style={edge, dashed, gray!90},
        reachability/.style={circle, fill opacity=0.1, draw opacity=0.3, dashed}
    ]

        \node[root] (s) at (0,0) {$s$};
        \node[task] (u) at (2.2,0) {$u$};
        \node[task] (d1) at (1.6,1.5) {$v$}; 
        
        \draw[edge] (s) -- (u) node[midway, below=1pt, font=\scriptsize] {$\psi(s_s)$};
        \draw[edge] (u) -- (d1) node[midway, left=2pt, font=\scriptsize] {$\psi(s_u)$};

        \node[task, fill=violet!5] (v2) at (4.4, 1.1) {$w$};
        \draw[candidate] (u) -- (v2) 
            node[pos=0.35, sloped, above, font=\tiny] {Small $\phi_\tau$}
            node[pos=0.75, sloped, below, font=\tiny, text=violet!80] {Small gain};
        
        \draw[reachability, fill=violet!80, draw=violet!80] (v2) circle (1.5cm);
        \node[future] (f1) at (4.6, 1.7) {}; 

        \node[task, fill=teal!5, thick] (v3) at (4.8,-1.5) {$z$};
        \draw[edge, teal] (u) -- (v3) 
            node[pos=0.35, sloped, above, font=\tiny, text=gray!60] {Large $\phi_{\tau}$}
            node[pos=0.75, sloped, below, font=\tiny] {Large $\mathrm{gain}$};
        
        \draw[reachability, fill=teal!80, draw=teal!80] (v3) circle (1.5cm);
        
        \node[future] (f2) at (5.3, -2.5) {};
        \node[future] (f3) at (5.6, -0.9) {}; 
        \node[future] (f4) at (4.1, -2.3) {};
        \node[future] (f5) at (5.7, -2.1) {}; 
				\node[future] (f6) at (4.8, -0.2) {};
				
				\node[future] (f7) at (6.6, -0.4) {}; 
				\node[future] (f8) at (3.0, -2.6) {}; 
				\node[future] (f9) at (6.5, -2.1) {}; 

    \end{tikzpicture}
\end{minipage}%
\hfill
\begin{minipage}[c]{0.45\linewidth}
    \caption{\textbf{CTL-MSTc tree construction strategy.} Supposing a cascade started from the root $s$ and has visited $u$ and $v$. In its next step, it needs to select one of the candidates ${w, z}$ to extend the tree. The strategy examines three factors for each candidate: (i) a penalty ($\psi$) that discourages extending the tree from $u$ 
if the path till there is already long; (ii) the cost to reach the new node ($\phi_\tau$) from an already visited node $u$, and (iii) the reach gain of the new node, related to the number of unvisited nodes (in gray) that are in its close vicinity. Here, despite $w$ being closer to $u$, node $z$ would be valued higher due to its higher gain.
		}
		\label{fig:ctl_mstc}
\end{minipage}
\end{figure}

We seek rooted cascades that jointly enforce transfer locality, structural balance, and global coverage. Greedy MST construction on raw distances $d(u,v)$ enforces locality but ignores two cascade-specific failure modes:
\textbf{(i)} transfer distances should not be penalized uniformly across scales, since nearby tasks often behave similarly whereas long-range transfers are substantially less reliable; moreover, nearby tasks may still connect to different regions of the graph, motivating a nonlinear distance shaping, and \textbf{(ii)} repeatedly attaching tasks to the same subtree creates deep paths that compound transfer error along the cascade. 
To address this, we introduce a locality penalty $\phi_{\tau}:\mathbb{R}_{+}\to\mathbb{R}_{+}$, parameterized by a scale $\tau$, encoding the distance threshold of Theorem~\ref{thm:ctl-vs-star}, and a subtree penalty $\psi:\mathbb{N}\to\mathbb{R}_{+}$ discouraging attachment to nodes with large existing subtrees, thereby promoting structural balance (specific choices are discussed in Appendix~\ref{app:budget_ablation}). The \emph{connection cost} of a partial tree $\mathcal{T}$ is then defined as:%
\begin{align*}
C(\mathcal{T}) = \sum_{w \notin \mathcal{T}} \min_{u \in \mathcal{T}} \phi_\tau\!\big(d(u,w)\big),
\end{align*}
i.e. the total penalty cost of connecting every remaining task to its nearest tree member. The \emph{reach gain}, denoted by $\mathrm{gain}(v\mid\mathcal{T})=C(\mathcal{T})-C(\mathcal{T}\cup\{v\})$, quantifies how much adding $v$ reduces future connection costs. Each candidate attachment $(u\!\to\!v)$ is scored by:%
\begin{align}
J(u,v\mid\mathcal{T})
= \psi\big(s_u(\mathcal{T})\big)\cdot\phi_{\tau}\big(d(u,v)\big)
  - \lambda\;\mathrm{gain}(v\mid\mathcal{T}),
\label{eq:mstc-score}
\end{align}
where $s_u(\mathcal{T})$ is the subtree size of $u$ in the current tree $\mathcal{T}$, and we iteratively select the edge minimizing $J$ over $u\in\mathcal{T}$, $v\notin\mathcal{T}$. $\mathrm{gain}(v\mid\mathcal{T})$ rewards candidates that reduce future connection cost for remaining tasks. As illustrated in Figure~\ref{fig:ctl_mstc}, this logic allows the strategy to bypass a `close' candidate (high locality but low gain) in favor of a more distant one that offers superior plain reachability. Setting $\lambda=0$, reduces to a greedy tree weighted by locality only; $\lambda>0$ introduces a plain coverage bias. We refer to this construction as \textsc{CTL-MSTc}.

\section{Experiments}
\label{sec:exps}

We evaluate CTL across synthetic and real-world regression and image classification tasks. 
In all experiments, task interactions occur exclusively through parameter initialization: each node is initialized by its parent and then refined locally. All models are linear (ridge regression for regression tasks, logistic regression with a linear head for classification), so the contraction assumption of Section~\ref{sec:ctl-theory} holds exactly throughout.

\subsection{Experimental Setup}

\paragraph{Synthetic datasets.}%
Linear regression tasks $\by_v = \bX_v\bT_v + \bE_v$ with i.i.d.\ Gaussian features share a $K$-cluster structure; the pair $(\tau_{\mathrm{within}},\tau_{\mathrm{between}})$ interpolates from nearly homogeneous to well-separated regimes. Full generative model and visualizations are in Appendix~\ref{app:synthetic}.

\paragraph{Real datasets.}%
For time-series regression, we use the WEAVE-UK electricity dataset\footnote{\url{https://weave.energy/}}: $500$ nodes in the Oxford area predicting feeder-level half-hourly demand from calendar and lagged-load features \citep{antoniadis2024hierarchical}. For image classification, we derive $200$ tasks from Fashion-MNIST \citep{xiao2017fashion} and CIFAR-$10$ \citep{krizhevsky2009learning} by sampling class pairs with replacement from the $\binom{10}{2}=45$ unique pairs (logistic regression on PCA embeddings), and $400$ tasks from CelebA \citep{liu2015faceattributes} drawn as a random subset of the $\binom{40}{2}=780$ unique binary attribute pairs (logistic regression on frozen ResNet-18 features). Architecture and preprocessing details are in Appendix~\ref{app:image_classif}.

\paragraph{Additional details.}%
All datasets use disjoint train/test splits. Synthetic tasks: $n_{\mathrm{train}}=64$, $n_{\mathrm{test}}=128$. WEAVE-UK electricity: training Feb 14–24, 2024; test Feb 25–28, 2024 (standard short-term load forecasting window \citep{antoniadis2024hierarchical}). Image tasks: fixed-size splits, same architecture across methods. Task graphs are built from training data only. The global budget $B$ is distributed via the SLE allocation formula across all experiments. Results are averaged over $20$ seeds. We perform one-sided Welch $t$-tests for two comparisons: CTL tree methods vs.\ Star (not shown in table, all significant for CTL-MST/MSTc), and CTL tree methods vs.\ Chain ($^\dagger$ in Table~\ref{tab:simplified-opt} indicates when $p<0.05$). A distance-proxy ablation is in Appendix~\ref{app:additional_results}; timing figures in Appendix~\ref{app:diagnostics}.

\subsection{Baselines}

All methods use identical training routines. We compare:
\textbf{Individual}: independent training per task (no transfer).
\textbf{Star}: all tasks initialized from a single seed task.
\textbf{Random Tree}: a random spanning tree (Prüfer sequence) rooted at the seed \citep{kumar1998parallel,deo2001prufer}; an uninformed cascade baseline.
\textbf{Chain}: greedy nearest-neighbor Hamiltonian path from the seed; strict locality but no branching, so, theoretically, error accumulates along the path.
\textbf{CTL-kNN}: BFS spanning tree through the $k$-NN graph ($k$ tuned per dataset); shallower and more branching than Chain.
\textbf{CTL-MST}: MST under the gradient distance, rooted at the medoid; the unpenalized CTL reference.
\textbf{CTL-MSTc}: the cascade-aware construction of Section~\ref{sec:mstc}, scored by~\eqref{eq:mstc-score}; balances locality and global reach. All CTL variants use the gradient distance, so performance differences isolate the effect of tree construction.

\subsection{Results}

Table~\ref{tab:simplified-opt} reports results across all settings (Figure~\ref{fig:radar} in the appendix summarizes relative gains over Star). Beyond the four baselines (Individual, Star, RandTree, Chain), we compare three CTL tree variants (kNN, MST, MSTc) all using the gradient distance.

\begin{table}[t]
\centering
\caption{Test performance across datasets and budgets $B$ (20 seeds; $\pm$ standard error). Synthetic regression reports MSE; WEAVE-UK and image classification report RMSE and accuracy respectively. Gray scales indicate top 3 methods (dark to light). \textbf{Bold} and \underline{underlined} denote best and second-best. $^\dagger$ marks CTL tree methods that significantly outperform Chain (one-sided Welch $t$-test, $p<0.05$).}
\label{tab:simplified-opt}
\resizebox{\textwidth}{!}{%
\begin{tabular}{ccc|cccc|ccc}
\toprule
\multicolumn{3}{c|}{Params}
& \multicolumn{4}{c|}{Baselines}
& \multicolumn{3}{c}{CTL} \\
Dataset & $T$ & $B$
& Indiv & Star & RandTree & Chain
& MST-kNN & MST & MSTc \\
\midrule
\multicolumn{10}{c}{\textbf{Synthetic Regression (MSE)}} \\
\midrule
Syn-$10$ & 200 & 500 & $948.9 \pm 33.8$ & $908.3 \pm 43.9$ & $904.9 \pm 53.1$ & \cellcolor{gray!20}$820.6 \pm 35.4$ & $875.2 \pm 51.1$ & \cellcolor{gray!60}$\mathbf{781.6 \pm 41.3}$ & \cellcolor{gray!40}$\underline{786.5 \pm 33.3}$ \\
Syn-$10$ & 200 & 1000 & $849.3 \pm 32.2$ & $802.2 \pm 46.4$ & \cellcolor{gray!20}$716.4 \pm 39.9$ & $658.3 \pm 29.8$ & $789.3 \pm 53.0$ & \cellcolor{gray!40}$\underline{642.6 \pm 32.1}$ & \cellcolor{gray!60}$\mathbf{606.0 \pm 25.1}$ \\
Syn-$10$ & 200 & 2000 & $910.9 \pm 47.0$ & $861.3 \pm 60.4$ & $788.0 \pm 41.0$ & $588.2 \pm 25.6$ & \cellcolor{gray!20}$723.3 \pm 35.9$ & \cellcolor{gray!60}$\mathbf{513.3 \pm 23.7}^\dagger$ & \cellcolor{gray!40}$\underline{574.2 \pm 26.8}$ \\
\midrule
\multicolumn{10}{c}{\textbf{WEAVE-UK Electricity Forecasting (RMSE, Wh)}} \\
\midrule
WEAVE-UK & $500$ & $500$ & $2521.4 \pm 1.0$ & $2024.4 \pm 12.3$ & $2039.8 \pm 16.1$ & \cellcolor{gray!20}$1474.2 \pm 27.2$ & $1908.6 \pm 59.6$ & \cellcolor{gray!40}$\underline{1450.2 \pm 18.5}$ & \cellcolor{gray!60}$\mathbf{1345.6 \pm 13.1}^\dagger$ \\
WEAVE-UK & $500$ & $1000$ & $2518.3 \pm 1.1$ & $1738.4 \pm 12.7$ & $1632.6 \pm 7.2$ & $1306.4 \pm 8.3$ & \cellcolor{gray!20}$1609.4 \pm 7.0$ & \cellcolor{gray!40}$\underline{1298.9\pm 7.0}$ & \cellcolor{gray!60}$\mathbf{1232.0 \pm 4.5}^\dagger$ \\
WEAVE-UK & $500$ & $2000$ & $2490.2 \pm 1.0$ & $1423.6 \pm 3.0$ & $1350.0 \pm 2.3$ & \cellcolor{gray!40}$\underline{1202.8 \pm 3.2}$ & $1365.0 \pm 11.6$ & \cellcolor{gray!20}$1209.8 \pm 2.8$ & \cellcolor{gray!60}$\mathbf{1168.0 \pm 1.8}^\dagger$ \\
\midrule
\multicolumn{10}{c}{\textbf{Image Classification (Accuracy, \%)}} \\
\midrule
FMNIST & $200$ & $500$ & $76.7 \pm 2.2$ & $77.6 \pm 2.0$ & $82.2 \pm 1.8$ & \cellcolor{gray!20}$93.1 \pm 1.1$ & $91.2 \pm 1.6$ & \cellcolor{gray!40}$\underline{93.1 \pm 1.2}$ & \cellcolor{gray!60}$\mathbf{93.3 \pm 1.0}$ \\
FMNIST & $200$ & $2000$ & $93.6 \pm 1.2$ & $93.3 \pm 1.1$ & $94.0 \pm 1.0$ & \cellcolor{gray!40}$\underline{95.7 \pm 0.9}$ & $93.3 \pm 1.1$ & \cellcolor{gray!20}$95.1 \pm 1.0$ & \cellcolor{gray!60}$\mathbf{96.0 \pm 0.8}$ \\
CIFAR & $200$ & $500$ & $53.9 \pm 0.3$ & $55.4 \pm 1.1$ & $57.0 \pm 1.0$ & $63.4 \pm 0.6$ & \cellcolor{gray!20}$63.6 \pm 1.0$ & \cellcolor{gray!60}$\mathbf{65.7 \pm 0.8}^\dagger$ & \cellcolor{gray!40}$\underline{63.8 \pm 0.7}$ \\
CIFAR & $200$ & $2000$ & $60.5 \pm 0.8$ & $57.0 \pm 1.1$ & $59.3 \pm 0.9$ & $66.9 \pm 0.7$ & \cellcolor{gray!40}$\underline{67.8 \pm 1.2}$ & \cellcolor{gray!60}$\mathbf{69.1 \pm 1.1}^\dagger$ & \cellcolor{gray!20}$67.3 \pm 1.1$ \\
CelebA & $400$ & $2000$ & $57.9 \pm 0.7$ & $59.6 \pm 1.7$ & $62.8 \pm 1.9$ & \cellcolor{gray!20}$77.6 \pm 1.5$ & \cellcolor{gray!60}$\mathbf{77.8 \pm 1.6}$ & \cellcolor{gray!40}$\underline{77.7 \pm 1.3}$ & $76.5 \pm 1.5$ \\
CelebA & $400$ & $4000$ & $55.9 \pm 1.2$ & $61.8 \pm 1.6$ & $65.4 \pm 1.5$ & \cellcolor{gray!60}$\mathbf{78.9 \pm 1.3}$ & \cellcolor{gray!40}$\underline{78.9 \pm 1.4}$ & $77.8 \pm 1.3$ & \cellcolor{gray!20}$77.9 \pm 1.4$ \\
\midrule
\multicolumn{3}{c|}{\textbf{Mean Rank}} & 6.92 & 6.08 & 5.17 & \cellcolor{gray!20}2.25 & 4.08 & \cellcolor{gray!40}\underline{1.83} & \cellcolor{gray!60}\textbf{1.67} \\
\bottomrule
\end{tabular}
}
\end{table}

CTL-MST and CTL-MSTc rank first and second overall (mean ranks 1.67 and 1.83), with consistent gains over Star across modalities: on WEAVE-UK at $B=500$, CTL-MSTc reduces RMSE by 46\% over Individual (33\% over Star); on CIFAR-10, CTL-MST gains +11.8 pp over Individual (+10.3 pp over Star). Gains are sharpest at low $B$, where warm-starting substitutes for early refinement steps, but persist as $B$ grows (18\% over Star on WEAVE-UK at $B=2000$), confirming CTL's value in resource-scarce regimes. 

Chain's performance is geometry-dependent: competitive on near-one-dimensional manifolds (FMNIST, CelebA) but dominated by tree methods in clustered geometries (WEAVE-UK, CIFAR, Syn-10), where six cells marked $^\dagger$ show statistical significance ($p<0.05$). CTL-MSTc outperforms CTL-MST in multi-cluster settings by steering attachment toward high-gain nodes, preventing deep single-branch paths that plain MST allows. MST-kNN underperforms despite identical distance metrics because BFS on $k$-NN graphs creates deeper paths; RandTree's marginal gains over Star confirm geometry-aware ordering is essential. 
The gradient proxy outperforms alternatives in all ablations (Appendix~\ref{app:additional_results}), as CTL transfers parameters and gradients directly capture optimization landscape proximity.

\section{Conclusion}
We introduced \emph{Cascaded Transfer Learning}, a framework for many-task learning under a strict global budget. Rather than independently adapting a shared source to every target, CTL propagates knowledge through a rooted tree over tasks, refining each task exactly once from its nearest trained neighbor. This single-pass design achieves frugality by construction: the total training cost is proportional to $B$, not to the number of tasks or their pairwise interactions. The theoretical analysis establishes that tree-structured cascades reduce transfer bias relative to direct transfer when the cascade is locally constructed and task distances are small relative to source-to-target distances. A key finding, absent from star-transfer and chain-based analyses, is that errors introduced early in the cascade are attenuated by every downstream refinement, so the effective cost of a long transfer is decomposed into a series of short, locally corrected steps. We further showed that trees outperform Hamiltonian chains at scale, because chains force linear depth growth and expose terminal tasks to long inter-cluster hops without downstream correction.
Empirically, CTL with gradient-based MST construction consistently and significantly improves over independent training, star transfer, and random trees across regression and classification tasks. The cascade-aware variant CTL-MSTc further surpasses the plain MST in multi-cluster settings, while the two are competitive where task geometry is approximately one-dimensional. These results suggest that structured, budget-aware transfer is a viable and scalable alternative to joint multi-task optimization in the many-task regime.

\section{Limitations and Future Work}
\label{sec:limitations}

\paragraph{Theoretical assumptions.}
The contraction assumption holds under strong convexity or the Polyak-Łojasiewicz condition, but is not guaranteed for deep networks with non-convex landscapes, where local refinement may converge to different basins. Restricting cascaded transfer to adapter layers or LoRA \citep{houlsby2019parameter,hu2022lora} mitigates this: freezing the backbone and transferring only a low-dimensional block keeps the restricted loss approximately strongly convex with respect to the effective weight update near the pre-trained initialization. Cascade depth and adapter rank then jointly control the error accumulation rate, offering a concrete handle on the theory-practice tradeoff. Appendix~\ref{app:pretrained} (Tables~\ref{tab:resnet} and~\ref{tab:chronos}) provides an empirical exploration of this direction: CTL improves over Star and Individual on top of frozen ResNet-18 and Chronos T5-small features, but in the settings we evaluated, plain linear CTL on raw features matches or exceeds these results at a fraction of the compute cost, suggesting that representation quality is not the binding constraint there.

\paragraph{Scalability and structural limitations.}
The $\mathcal{O}(T^2)$ cost of pairwise distance computation is the dominant bottleneck for large $T$; approximate nearest-neighbor methods \citep{malkov2018efficient} on gradient embeddings could reduce this to $\mathcal{O}(T\log T)$. MSTs do not control worst-case path length: depth-error correlations remain near zero empirically (Appendix~\ref{app:diagnostics}), and CTL outperforms independent training at every depth level, but minimum-bottleneck spanning trees (which minimize the maximum edge weight rather than total path length) and cascade forests over clustered tasks are natural directions that CTL-MSTc heuristically encourages but does not formally subsume. Proxy quality may degrade for heterogeneous architectures; fine-tuning-based probing or model mergeability scores are natural extensions. Relaxing the tree structure to DAGs would further enable multi-parent knowledge fusion.

\bibliographystyle{plain}
\bibliography{main}%

\appendix

\section*{Broader Impact}

This work advances Machine Learning by studying scalable transfer mechanisms across large collections of related tasks. The proposed Cascaded Transfer Learning framework emphasizes computational frugality: tasks are refined locally under an explicit global budget, avoiding repeated joint optimization or large-scale retraining. By favoring lightweight, localized updates over centralized training, CTL reduces redundant computation and can lower energy consumption in many-task and distributed learning settings.

\section{Code Availability}

To promote transparency and reproducibility, we provide an implementation of Cascaded Transfer Learning together with scripts to reproduce the experiments reported in this paper. The code and documentation are available through the project website at
\href{https://cascader-ctl.github.io/CASCADER/}{this address}.

\section{Computational Details}
\label{app:compute}

All experiments were run on a single Apple M4 Pro laptop (ARM architecture, unified memory). No external GPU or cloud compute was used. For the main CTL variants (linear models, $T \le 200$ tasks), a full 20-seed sweep over three budget values completes in under 5 minutes. Graph construction (MST/MSTc) accounts for less than 1\% of total training time. The pretrained-feature experiments are more expensive: LoRA fine-tuning on WEAVE-UK takes $>$8 min per run, versus ${\sim}30$ s for the linear CTL baseline on raw features.

\section{Proofs for Theoretical Analysis of Cascaded Transfer Learning}
\label{app:tree_proofs}

\subsection{Parameter-Space}

\begin{proof}[Cascaded transfer over a path]
By definition of the CTL update, $\tilde\bT_v = G_v^{b_v}(\tilde\bT_{\mathrm{pa}(v)})$. By the contractive refinement assumption,
\begin{equation*}
    \bigl\lVert\tilde\bT_v-\bT_v^\star\bigr\rVert=\bigl\lVert G_v^{b_v}(\tilde\bT_{\mathrm{pa}(v)})-\bT_v^\star\bigr\rVert \le \rho_v^{b_v}\bigl\lVert\tilde\bT_{\mathrm{pa}(v)}-\bT_v^\star\bigr\rVert.
\end{equation*}
Applying the triangle inequality,
\begin{equation*}
    \bigl\lVert\tilde\bT_{\mathrm{pa}(v)}-\bT_v^\star\bigr\rVert\le\bigl\lVert\tilde\bT_{\mathrm{pa}(v)}-\bT_{\mathrm{pa}(v)}^\star\bigr\rVert + d(pa(v),v).
\end{equation*}
Combining the two inequalities yields the result.
\end{proof}

\begin{proof}[Path-wise error decomposition]
Applying the above to node $v_1$:
\begin{equation*}
    \bigl\lVert\tilde\bT_{v_1}-\bT_{v_1}^\star\bigr\rVert\le\rho_{v_1}^{b_{v_1}}\Big(\bigl\lVert\tilde\bT_{v_0}-\bT_{v_0}^\star\bigr\rVert+ d(v_0,v_1)\Big).
\end{equation*}
Same with node $v_2$ and substitute the previous inequality:
\begin{equation*}
    \bigl\lVert\tilde\bT_{v_2}-\bT_{v_2}^\star\bigr\rVert\le\rho_{v_1}^{b_{v_1}}\rho_{v_2}^{b_{v_2}}\bigl\lVert\tilde\bT_{v_0}-\bT_{v_0}^\star\bigr\rVert+\rho_{v_1}^{b_{v_1}}\rho_{v_2}^{b_{v_2}} d(v_0,v_1)+\rho_{v_2}^{b_{v_2}} d(v_1,v_2).
\end{equation*}
Proceeding inductively, each application introduces a multiplicative factor $\rho_{v_i}^{b_{v_i}}$ on all upstream terms and adds a new edge term. Therefore, we introduce a multiplicative notation, for $i,m\in\mathbb N^\star, P_{i:m}= \prod_{j=i}^m\rho_{v_j}^{b_{v_j}}$.  Hence, we get 
\begin{align*}
    \bigl\lVert\tilde\bT_{v}-\bT_{v}^\star\bigr\rVert\le P_{1:m}\bigl\lVert\tilde\bT_{v_0}-\bT_{v_0}^\star\bigr\rVert+\sum_{i=1}^m P_{i:m} d(v_{i-1},v_i).
\end{align*}
Setting $\tilde\bT_{v_0} = \bT_{v_0}^\star$ cancels out the bias term, enforcing $b_{v_i} = b$ for all $1\le i \le m$, plus bounding all $\rho_{v_i}$ by $\rho_{\max}$ and $\delta_i$ by $\delta_{\max}$ yields a finite geometric series of finite ratio $\rho_{\max}^b$.
\end{proof}

\subsection{Feature-Space}

\begin{proof}[Feature-space contraction]
The gradient update is $G_v(\bT)=\bT-\eta \bX_v^\top(\bX_v\bT-\by_v)= \bM_v\bT+\eta \bX_v^\top \by_v$. Since $\bT_v^\star$ is the OLS solution, it satisfies the normal equations $\bX_v^\top \bX_v\bT_v^\star=\bX_v^\top \by_v$ by definition. Therefore $G_v(\bT_v^\star)=\bM_v\bT_v^\star+\eta\bX_v^\top\by_v=\bT_v^\star$,
so $\bT_v^\star$ is a fixed point. Thus, $G_v(\bT)-\bT_v^\star=\bM_v(\bT-\bT_v^\star)$. Iterating yields the claim.
\end{proof}

\begin{proof}[Edge-wise propagation in feature space]
The above yields $\tilde\bT_v-\bT_v^\star=\bM_v^{b_v}(\tilde\bT_{\mathrm{pa}(v)}-\bT_v^\star)$. Taking norms and using $\bigl\lVert\bM_v^{b_v}\bigr\rVert\le\rho_v^{b_v}$,
\begin{equation*}
    \bigl\lVert\tilde\bT_v-\bT_v^\star\bigr\rVert\le\rho_v^{b_v}\bigl\lVert\tilde\bT_{\mathrm{pa}(v)}-\bT_v^\star\bigr\rVert.
\end{equation*}
The result follows by the triangle inequality exactly as in the parameter-space case.
\end{proof}

\begin{proof}[Path-wise error propagation in feature space]
The argument follows verbatim from the parameter-space analysis. Unrolling the edge-wise inequality along the unique root-to-$v$ path yields the stated bound, with each edge contribution geometrically damped by downstream refinements.
\end{proof}

\subsection{Noisy Feature-Space}

\begin{proposition}[Expected noisy propagation along a path]
\label{prop:noisy_propagation}
Under the noisy linear model ($\by_v = \bX_v\bT_v^\star + \boldsymbol\varepsilon_v$, mean-zero sub-Gaussian noise with independent coordinates and variance $\sigma_v^2$ per coordinate) and the feature-space contraction assumption, assuming $\bX_v^\top\bX_v\succ 0$ (equivalently, replace $\bX_v^\top\bX_v$ by $\bX_v^\top\bX_v+\lambda\bId$ for ridge, which is always positive definite), using the notation of Proposition~\ref{prop:propagation-param}:
\begin{align*}
\mathbb{E}\bigl[\bigl\lVert\tilde\bT_v-\bT_v^\star\bigr\rVert\bigr]
\le P_{1:m}\,\mathbb{E}\bigl[\bigl\lVert\tilde\bT_s-\bT_s^\star\bigr\rVert\bigr]
+\sum_{i=1}^m P_{i:m}\,d(v_{i-1},v_i)
+\sum_{i=1}^m P_{i+1:m}\,\sigma_{v_i}\bigl(1+\rho_{v_i}^{b_{v_i}}\bigr)\|\bA_{v_i}\|_F,
\end{align*}
where $\bA_v=(\bX_v^\top\bX_v)^{-1}\bX_v^\top$. Locality reduces transfer bias; sufficient downstream refinement limits noise accumulation.
\end{proposition}

\begin{proof}[Empirical optimum]
The minimizer $\hat\bT_v$ of $\tfrac12\bigl\|\bX_v\bT-\by_v\bigr\|^2$ satisfies the equations
\begin{equation*}
    \bX_v^\top\bX_v\hat\bT_v=\bX_v^\top\by_v=\bX_v^\top(\bX_v\bT_v^\star+\varepsilon_v) =\bX_v^\top\bX_v\,\bT_v^\star+\bX_v^\top\boldsymbol{\varepsilon}_v.
\end{equation*}
Since $\bX_v^\top\bX_v \succ 0$,
\begin{equation*}
    \hat\bT_v=\bT_v^\star+(\bX_v^\top\bX_v)^{-1}\bX^\top\boldsymbol{\varepsilon}_v.
\end{equation*}
Equivalently, under Gaussian noise, the estimator $\hat\bT_v$ is normally distributed with mean $\bT_v^\star$ and covariance $\sigma_v^2\bA_v\bA_v^\top$. Moreover, the gradient update is $G_v(\bT)=\bM_v\bT+\eta\bX^\top\by_v$,
with $\bM_v=\bId-\eta\bX_v^\top\bX_v$. Using $\bX_v^\top\by_v=\bX_v^\top\bX_v\,\hat\bT_v$, we have that $\hat\bT_v$ is a fixed point. Hence $G_v(\bT)-\hat\bT_v=\bM_v(\bT-\hat\bT_v)$, and iterating gives $G_v^{b_v}(\bT)-\hat\bT_v=\bM_v^{b_v}(\bT-\hat\bT_v)$.
\end{proof}

\begin{proof}[Expected estimation error]
Let $\bA_v=(\bX_v^\top\bX_v)^{-1}\bX_v^\top$. From the empirical optimum lemma, $\hat\bT_v-\bT_v^\star=\bA_v\bE_v$. Using Jensen's inequality yields $\mathbb E\big[\big\|\bA_v\bE_v\big\|\big]\le \sqrt{\mathbb E\big[\big\|\bA_v\bE_v\big\|^2\big]}$.
Now,
\begin{equation*}
    \bigl\lVert\bA_v\bE_v\bigr\rVert^2=\bE_v^\top \bA_v^\top\bA_v\bE_v =\sum_{i,j} (\bA_v^\top\bA_v)_{ij}\bE_{v,i}\bE_{v,j}.
\end{equation*}
Since the coordinates are independent and mean-zero, $\mathbb E[\bE_{v,i}\bE_{v,j}]=0$ for $i\neq j$. Therefore,
\begin{equation*}
    \mathbb E\big[\bigl\lVert\bA_v\bE_v\bigr\rVert^2\big]=\sum_i (\bA_v^\top\bA_v)_{ii}\mathbb E[\bE_{v,i}^2]\le \sigma_v^2 \sum_i (\bA_v^\top\bA_v)_{ii}=\sigma_v^2\mathrm{tr}(\bA_v^\top\bA_v)=\sigma_v^2\|\bA_v\|_F^2.
\end{equation*}
Combining yields $\mathbb E[\bigl\lVert\hat\bT_v-\bT_v^\star\bigr\rVert]\le \sigma_v\|\bA_v\|_F$. 
\end{proof}

\begin{proof}[Expected noisy edge-wise propagation]
With the contraction assumption for task $v$, we get
\begin{align*}
    \bigl\lVert\tilde\bT_v-\hat\bT_v\bigr\rVert\le \rho_v^{b_v}\bigl\lVert\tilde\bT_u-\hat\bT_v\bigr\rVert\,.
\end{align*}
Then, using the triangle inequality,
\begin{align*}
    \bigl\lVert\tilde\bT_u-\hat\bT_v\bigr\rVert\le \bigl\lVert\tilde\bT_u-\bT_v^\star\bigr\rVert+\bigl\lVert\hat\bT_v-\bT_v^\star\bigr\rVert\le \bigl\lVert\tilde\bT_u-\bT_u^\star\bigr\rVert+\bigl\lVert\bT_u^\star-\bT_v^\star\bigr\rVert+\bigl\lVert\hat\bT_v-\bT_v^\star\bigr\rVert\,.
\end{align*}
Therefore,
\begin{align*}
    \bigl\lVert\tilde\bT_v-\bT_v^\star\bigr\rVert\le\bigl\lVert\tilde\bT_v-\hat\bT_v\bigr\rVert+\bigl\lVert\hat\bT_v-\bT_v^\star\bigr\rVert\le\rho_v^{b_v}\big(\bigl\lVert\tilde\bT_u-\bT_u^\star\bigr\rVert+\bigl\lVert\bT_u^\star-\bT_v^\star\bigr\rVert+\bigl\lVert\hat\bT_v-\bT_v^\star\bigr\rVert\big)+\bigl\lVert\hat\bT_v-\bT_v^\star\bigr\rVert\,.
\end{align*}
Taking expectations and using $\mathbb E\left[\bigl\lVert\hat\bT_v-\bT_v^\star\bigr\rVert\right]\le \sigma_v\|\bA_v\|_F$ gives
\begin{align*}
    \mathbb E\left[\bigl\lVert\tilde\bT_v-\bT_v^\star\bigr\rVert\right]\le\rho_v^{b_v}\Big(\mathbb E\left[\bigl\lVert\tilde\bT_u-\bT_u^\star\bigr\rVert\right]+\bigl\lVert\bT_u^\star-\bT_v^\star\bigr\rVert\Big)+\sigma_v(1+\rho_v^{b_v})\|\bA_v\|_F\,,
\end{align*}
as claimed.
\end{proof}

\begin{proof}[Expected path-wise propagation]
By the expected noisy edge-wise propagation lemma, for each $i\ge 1$,
\begin{equation*}
  \mathbb E\big[\bigl\lVert\tilde\bT_{v_i}-\bT_{v_i}^\star\bigr\rVert\big] \le \rho_{v_i}^{b_{v_i}}\big(\mathbb E\big[\bigl\lVert\tilde\bT_{v_{i-1}}-\bT_{v_{i-1}}^\star\bigr\rVert\big]+\bigl\lVert\bT_{v_{i-1}}^\star-\bT_{v_i}^\star\bigr\rVert\big)+\sigma_{v_i}\bigl(1+\rho_{v_i}^{b_{v_i}}\bigr)\|\bA_{v_i}\|_F\,.
\end{equation*}
Applying this inequality to $v_1$ yields
\begin{align*}
    \mathbb E\big[\bigl\lVert\tilde\bT_{v_1}-\bT_{v_1}^\star\bigr\rVert\big] \le \rho_{v_1}^{b_{v_1}}\Big(\mathbb E\big[\bigl\lVert\tilde\bT_{v_0}-\bT_{v_0}^\star\bigr\rVert\big]+\bigl\lVert\bT_{v_0}^\star-\bT_{v_1}^\star\bigr\rVert\Big)+\mathcal \sigma_{v_1}\bigl(1+\rho_{v_1}^{b_{v_1}}\bigr)\|\bA_{v_1}\|_F\,.
\end{align*}
Applying the same inequality to $v_2$ and bounding the expectation term,
\begin{align*}
    \mathbb E\big[\bigl\lVert\tilde\bT_{v_2}-\bT_{v_2}^\star\bigr\rVert\big] \le \rho_{v_2}^{b_{v_2}}\rho_{v_1}^{b_{v_1}} \mathbb E\big[\bigl\lVert\tilde\bT_{v_0}-\bT_{v_0}^\star\bigr\rVert\big]
+ \rho_{v_2}^{b_{v_2}}\rho_{v_1}^{b_{v_1}}\bigl\lVert\bT_s^\star-\bT_{v_1}^\star\bigr\rVert
&+ \rho_{v_2}^{b_{v_2}}\bigl\lVert\bT_{v_1}^\star-\bT_{v_2}^\star\bigr\rVert\\
&+ \rho_{v_2}^{b_{v_2}}\sigma_{v_1}\bigl(1+\rho_{v_1}^{b_{v_1}}\bigr)\|\bA_{v_1}\|_F\\
&+\sigma_{v_2}\bigl(1+\rho_{v_2}^{b_{v_2}}\bigr)\|\bA_{v_2}\|_F\,.
\end{align*}
Proceeding inductively, each step multiplies all upstream terms by $\rho_{v_i}^{b_{v_i}}$ and adds a new distance term discounted by downstream contractions, together with a new noise contribution discounted only by refinements performed after node $v_i$.  Collecting terms yields
\begin{align*}
    \mathbb E\big[\bigl\lVert\tilde\bT_{v}-\bT_{v}^\star\bigr\rVert\big]\le P_{1:m}\mathbb E\big[\bigl\lVert\tilde\bT_{v_0}-\bT_{v_0}^\star\bigr\rVert\big]&+\sum_{i=1}^m P_{i:m}\bigl\lVert\bT_{v_{i-1}}^\star-\bT_{v_i}^\star\bigr\rVert \\ 
    &+\sum_{i=1}^m P_{i+1:m}\sigma_{v_i}\bigl(1+\rho_{v_i}^{b_{v_i}}\bigr)\|\bA_{v_i}\|_F\,,
\end{align*}
with the convention that empty products equal $1$. This concludes the proof.
\end{proof}

\section{Distance metrics for MST-based cascades}

We group the task distance metrics used to construct MST-based cascades into the following families, summarized in Figure~\ref{fig:distance-taxonomy} according to the type of information they require and the level at which task similarity is assessed. Some of these measures are divergences rather than true metrics, but are used as nonnegative edge weights for MST construction.

\begin{itemize}[leftmargin=1.5em,itemsep=0.em,topsep=0.em]
\item \textbf{Feature-based distances.}
These distances compare task feature representations $\bX_v \in \RR^{n \times d}$.

\begin{itemize}[leftmargin=1.2em,itemsep=0.em]
    \item \emph{Feature distance.}
    When feature matrices have identical shape, we use a normalized Euclidean distance between flattened representations 
    \begin{align*}
        d_{\mathrm{feat}}(u,v)
        = \sqrt{\frac{1}{D} \Bigl\lVert \mathrm{vec}(\bX_u) - \mathrm{vec}(\bX_v) \Bigr\rVert^2}\,,
    \end{align*}
    where $D$ is the number of overlapping dimensions. When shapes differ, each task is embedded using concatenated empirical means and variances, and the same normalized Euclidean distance is applied.

    \item \emph{Gaussian mean--covariance distance.}
    Approximating each feature distribution by a Gaussian $\mathcal{N}(\boldsymbol \mu_v\,,\bS_v)$, we define
    \begin{align*}
        d_{\mathrm{Gauss}}(u,v)
        = \lVert\boldsymbol\mu_u - \boldsymbol\mu_v\rVert + \lVert\bS_u - \bS_v\rVert_F\,.
    \end{align*}
    
    \item \emph{Centered Kernel Alignment (CKA) distance.}
    Given centered representations $\bZ_u \in \RR^{n \times d}$ and $\bZ_v \in \RR^{n \times d}$, we use linear kernels $\bK_u=\bZ_u\bZ_u^\top$ and $\bK_v=\bZ_v\bZ_v^\top$ and their centered versions $\tilde{\bK}_u = \bH \bK_u \bH$\,, $\tilde \bK_v = \bH \bK_v \bH$\,, where $\bH = \bIn - \frac{1}{n}\mathbf 1\mathbf 1^\top$.
    We define
    \begin{align*}
        \mathrm{HSIC}(\bK_u,\,\bK_v) = \mathrm{tr}\bigl(\tilde \bK_u \tilde \bK_v\bigr)\,,
    \end{align*}
    where HSIC stands for Hilbert-Schmidt Independence Criterion, and define the CKA similarity as
    \begin{align*}
      \mathrm{CKA}(u,v)
    = \frac{\mathrm{HSIC}(\bK_u,\,\bK_v)}
    {\sqrt{\mathrm{HSIC}(\bK_u,\,\bK_u)\,\mathrm{HSIC}(\bK_v,\,\bK_v)}}\,.  
    \end{align*}
    Finally, we set the dissimilarity to $d_{\mathrm{CKA}}(u,v)=1-\mathrm{CKA}(u,v)$.
\end{itemize}

\item \textbf{Target-based distances.}
These distances compare task outputs $\by_v \in \RR^{n}$ directly.
\begin{itemize}[leftmargin=1.2em,itemsep=0.em]
    \item \emph{Target distance.}
    \begin{align*}
        d_{\mathrm{target}}(u,v)
        = \bigl\lVert \by_u - \by_v \bigr\rVert\,.
    \end{align*}

    \item \emph{Symmetric Kullback--Leibler (KL) divergence.}
    Let $p_v$ denote a smoothed empirical histogram of $\by_v$. We define
    \begin{align*}
        d_{\mathrm{KL}}(u,v)
        = \tfrac12 \bigl( \mathrm{KL}(p_u\|p_v) + \mathrm{KL}(p_v\|p_u) \bigr)\,.
    \end{align*}
    \item \emph{Jensen--Shannon distance.}
    \begin{align*}
        d_{\mathrm{JS}}(u,v)
        = \sqrt{\tfrac12 \mathrm{KL}(p_u\|  m) + \tfrac12 \mathrm{KL}(p_v\| m)}\,,
        \quad m=\tfrac12(p_u+p_v).
    \end{align*}

    \item \emph{Wasserstein distance.}
    We use the 1-Wasserstein distance between empirical target distributions $\hat p_u$ and $\hat p_v$,
    \begin{align*}
        d_{\mathrm{Wass}}(u,v)
        = W_1(\hat{p}_u,\,\hat{p}_v)\,.
    \end{align*}
\end{itemize}

\item \textbf{Optimization-geometry distances.}
These distances are derived from optimization quantities or model parameters.

\begin{itemize}[leftmargin=1.2em,itemsep=0.em]
    \item \emph{Gradient distance.}
    Using gradients at initialization $\bg_v = \bX_v^\top \by_v$, optionally normalized,
    \begin{align*}
        d_{\mathrm{grad}}(u,v)
        = \lVert \bg_u - \bg_v \rVert\,.
    \end{align*}
        
    \item \emph{Model distance.}
    Let $\hat\bT_v$ denote the ridge-regression solution for task $v$. We define
    \begin{align*}
        d_{\mathrm{model}}(u,v)
        = \lVert \hat\bT_u - \hat\bT_v \rVert\,.
    \end{align*}
\end{itemize}

\end{itemize}

\begin{figure}[H]
\centering
\small
\resizebox{\columnwidth}{!}{
\begin{tikzpicture}[
    node distance=8mm,
    rounded corners,
    font=\small,
    box/.style={draw, rounded corners, align=left, inner sep=6pt, line width=0.8pt, text width=67mm},
    infobox/.style={box, fill=blue!10, text width=48mm},
    methodbox/.style={box, fill=orange!15, text width=57mm},
    detailbox/.style={box, fill=yellow!10, text width=67mm, font=\footnotesize},
    arrow/.style={->, >=stealth, thick, draw=gray!70}
]

\node[infobox] (info) {
    \textbf{Available information}\\[2pt]
    \(\bullet\) Features \(\bX_v\)\\
    \(\bullet\) Targets \(\by_v\)\\
    \(\bullet\) Gradients at initialization\\
    \(\bullet\) Model parameters / solutions
};

\node[methodbox, right=10mm of info, yshift=25mm] (feature) {
    \textbf{Feature-based distances}\\[1pt]
    Compare task feature representations \(\bX_v\)
};
\node[methodbox, right=10mm of info, yshift=0mm] (target) {
    \textbf{Target-based distances}\\[1pt]
    Compare task outputs \(\by_v\) directly
};
\node[methodbox, right=10mm of info, yshift=-25mm] (optim) {
    \textbf{Optimization-geometry distances}\\[1pt]
    Derived from optimization/parameters
};

\draw[arrow] (info.east) -- (feature.west);
\draw[arrow] (info.east) -- (target.west);
\draw[arrow] (info.east) -- (optim.west);

\node[detailbox, right=10mm of feature] (featurev) {
    \textbf{Methods}\\[1pt]
    \(\circ\) Euclidean on flattened \(\bX_v\) (feature distance)\\
    \(\circ\) Mean-variance embedding (shape mismatch)\\
    \(\circ\) Gaussian approximation (means \& covariances)\\
    \(\circ\) Linear CKA (representation-level)
};
\node[detailbox, right=10mm of target] (targetv) {
    \textbf{Methods}\\[1pt]
    \(\circ\) Euclidean on flattened \(\by_v\) (target distance)\\
    \(\circ\) Symmetric KL (smoothed histograms)\\
    \(\circ\) Jensen-Shannon distance\\
    \(\circ\) $1$-Wasserstein / Earth Mover's distance
};
\node[detailbox, right=10mm of optim] (optimv) {
    \textbf{Methods}\\[1pt]
    \(\circ\) Gradient proxy \(\bg_v = \bX_v^\top \by_v\) (gradient distance)\\
    \(\circ\) Ridge-regression solutions (model distance)
};

\draw[arrow] (feature.east) -- (featurev.west);
\draw[arrow] (target.east) -- (targetv.west);
\draw[arrow] (optim.east) -- (optimv.west);

\end{tikzpicture}}
\caption{Distance taxonomy for MST-based cascade construction.}
\label{fig:distance-taxonomy}
\end{figure}

Among these families, the optimization-geometry distances perform most consistently. A proxy-quality audit on synthetic regression tasks shows that the gradient proxy yields the largest average gain over Star ($+31.7\%$), while feature- and target-based proxies achieve more modest gains ($\approx17$--$18\%$). This motivates the gradient distance as the default choice throughout all experiments. The taxonomy is provided for completeness and to support practitioners working in settings where gradients are unavailable, for instance when tasks share a common unlabeled feature space but differ only in their annotation protocol.

\section{Additional Experimental Results}
\label{app:additional_results}

\begin{figure}[t]
    \centering
    \includegraphics[width=\linewidth]{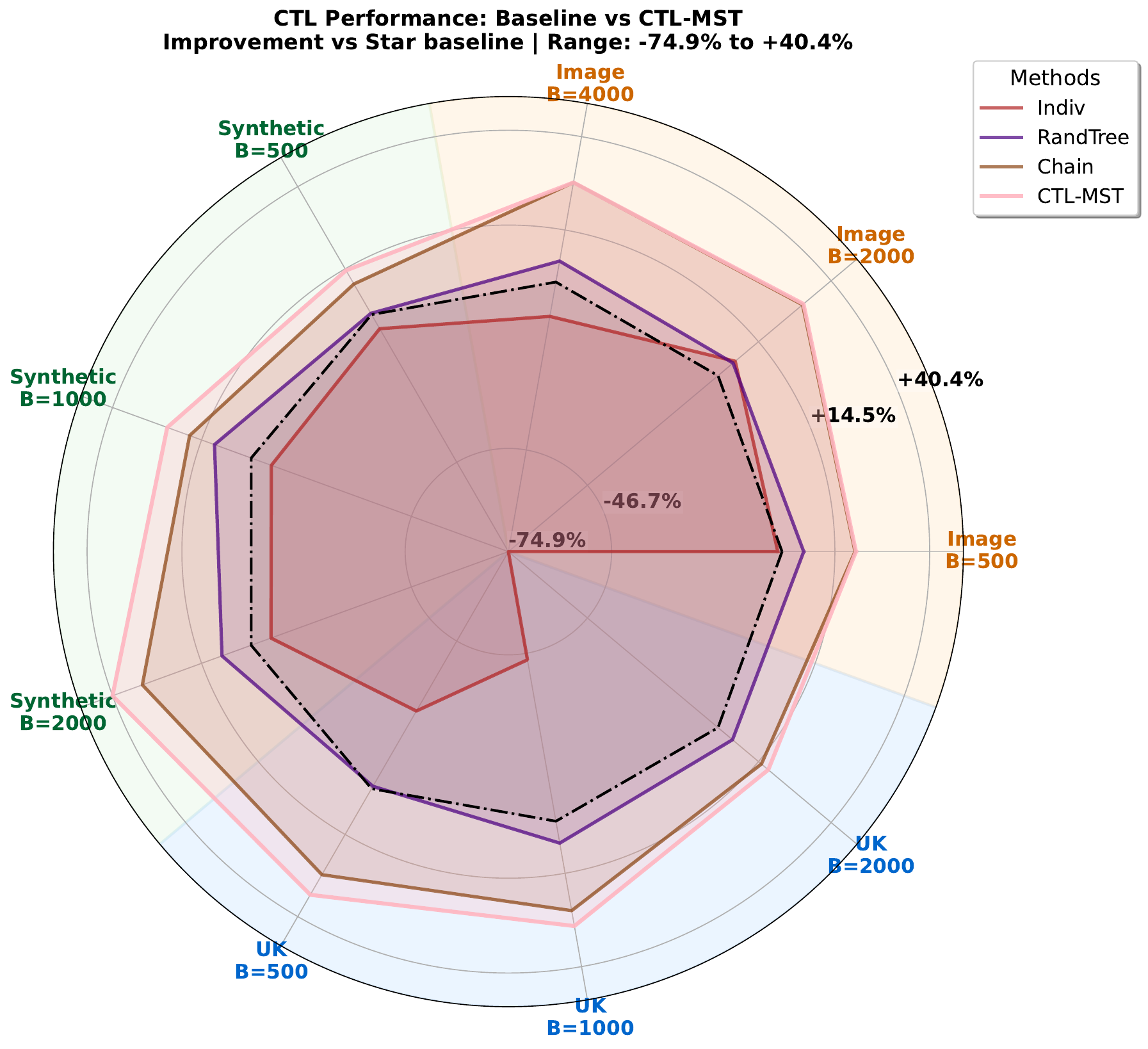}
    \caption{Radar plot of average improvement (\%) over the Star baseline, across datasets and budgets.}
    \label{fig:radar}
\end{figure}

This section provides an ablation study on the choice of distance proxy for MST construction. All CTL-MST variants use the same cascade algorithm; only the task distance used to build the MST differs. The gradient proxy is used as the reference.

\begin{table*}[t]
\centering
\caption{Distance proxy ablation for CTL-MST ($50$ seeds). Each entry is the signed performance gap of the given proxy relative to the gradient baseline. \textbf{Regression} ($\downarrow$; metric specified per row): $\Delta = (X_{\mathrm{metric}} - \mathrm{Grad}_{\mathrm{metric}})/\mathrm{Grad}_{\mathrm{metric}}\times 100\%$; positive = gradient wins (alternative worse). \textbf{Classification} (Acc $\uparrow$): $\Delta = \mathrm{Grad}_{\mathrm{acc}} - X_{\mathrm{acc}}$ in pp; positive = gradient wins (alternative worse). \textcolor{darkred}{\textbf{Dark red}}: way worse ($>10\%$ or $>5\,\text{pp}$). \textcolor{red}{\textbf{Red}}: worse. \textcolor{darkgreen}{\textbf{Green}}: better. \textcolor{deepgreen}{\textbf{Dark green}}: way better ($<{-5\%}$).}
\label{tab:all_experiments}
\resizebox{\textwidth}{!}{%
\begin{tabular}{ccc | ccc | cccc | c}
\toprule
& & & \multicolumn{3}{c|}{Feature-based ($\Delta$)} & \multicolumn{4}{c|}{Target-based ($\Delta$)} & Optim ($\Delta$) \\
Dataset & $T$ & $B$ & Feature & MeanCov & CKA & Target & KL & Wasserstein & JS & Model \\
\midrule
\multicolumn{11}{c}{\textbf{Synthetic Regression (MSE $\downarrow$; $\Delta$ in \%)}} \\
\midrule
Syn-2  & 200 & 500  & \textcolor{red}{$+3.6$}  & \textcolor{red}{$+6.6$}  & \textcolor{red}{$+6.0$}  & \textcolor{red}{$+3.0$}  & \textcolor{red}{$+1.8$}  & \textcolor{red}{$+4.9$}  & \textcolor{darkred}{$+12.3$} & \textcolor{deepgreen}{$-5.7$}  \\
Syn-2  & 200 & 1000 & \textcolor{darkgreen}{$-1.5$}  & \textcolor{darkgreen}{$-3.6$}  & \textcolor{red}{$+4.5$}  & \textcolor{red}{$+1.2$}  & \textcolor{red}{$+0.2$}  & \textcolor{darkred}{$+11.9$} & \textcolor{red}{$+3.8$}  & \textcolor{darkgreen}{$-3.2$}  \\
Syn-2  & 200 & 2000 & \textcolor{darkred}{$+21.3$} & \textcolor{darkred}{$+15.9$} & \textcolor{red}{$+9.1$}  & \textcolor{darkred}{$+13.1$} & \textcolor{darkred}{$+13.9$} & \textcolor{darkred}{$+17.0$} & \textcolor{darkred}{$+23.6$} & \textcolor{deepgreen}{$-6.1$}  \\
Syn-5  & 200 & 500  & \textcolor{red}{$+0.6$}  & \textcolor{red}{$+2.3$}  & \textcolor{red}{$+4.2$}  & \textcolor{darkgreen}{$-0.7$}  & \textcolor{red}{$+3.4$}  & \textcolor{red}{$+0.9$}  & \textcolor{red}{$+3.9$}  & \textcolor{red}{$+1.2$}  \\
Syn-5  & 200 & 1000 & \textcolor{red}{$+1.9$}  & \textcolor{red}{$+5.8$}  & \textcolor{red}{$+7.1$}  & \textcolor{red}{$+2.1$}  & \textcolor{red}{$+4.3$}  & \textcolor{red}{$+7.0$}  & \textcolor{darkgreen}{$-0.9$}  & \textcolor{red}{$+0.7$}  \\
Syn-5  & 200 & 2000 & \textcolor{darkred}{$+18.4$} & \textcolor{darkred}{$+18.3$} & \textcolor{darkred}{$+24.7$} & \textcolor{darkred}{$+16.1$} & \textcolor{darkred}{$+16.8$} & \textcolor{darkred}{$+18.4$} & \textcolor{darkred}{$+18.3$} & \textcolor{red}{$+2.2$}  \\
Syn-10 & 200 & 500  & \textcolor{darkgreen}{$-1.0$}  & \textcolor{red}{$+1.3$}  & \textcolor{red}{$+0.1$}  & \textcolor{red}{$+1.6$}  & \textcolor{red}{$+5.6$}  & \textcolor{red}{$+1.9$}  & \textcolor{deepgreen}{$-5.2$}  & \textcolor{darkgreen}{$-4.7$}  \\
Syn-10 & 200 & 1000 & \textcolor{red}{$+3.9$}  & \textcolor{red}{$+5.9$}  & \textcolor{red}{$+5.7$}  & \textcolor{red}{$+8.0$}  & \textcolor{red}{$+7.5$}  & \textcolor{red}{$+5.2$}  & \textcolor{red}{$+5.1$}  & \textcolor{deepgreen}{$-5.1$}  \\
Syn-10 & 200 & 2000 & \textcolor{darkred}{$+15.0$} & \textcolor{darkred}{$+18.8$} & \textcolor{darkred}{$+15.2$} & \textcolor{darkred}{$+12.7$} & \textcolor{darkred}{$+11.9$} & \textcolor{darkred}{$+14.0$} & \textcolor{red}{$+9.4$}  & \textcolor{darkgreen}{$-2.1$}  \\
\midrule
\multicolumn{11}{c}{\textbf{WEAVE-UK Electricity (RMSE $\downarrow$, Wh; $\Delta$ in \%)}} \\
\midrule
WEAVE-UK & 500 & 500  & \textcolor{darkred}{$+21.6$} & \textcolor{darkred}{$+21.9$} & \textcolor{darkred}{$+18.4$} & \textcolor{red}{$+8.3$}  & \textcolor{red}{$+2.6$}  & \textcolor{red}{$+2.1$}  & \textcolor{red}{$+6.2$}  & \textcolor{red}{$+5.0$}  \\
WEAVE-UK & 500 & 1000 & \textcolor{darkred}{$+20.7$} & \textcolor{darkred}{$+20.9$} & \textcolor{darkred}{$+21.7$} & \textcolor{red}{$+7.1$}  & \textcolor{red}{$+4.6$}  & \textcolor{darkgreen}{$-0.7$}  & \textcolor{red}{$+3.9$}  & \textcolor{red}{$+5.0$}  \\
WEAVE-UK & 500 & 2000 & \textcolor{darkred}{$+18.5$} & \textcolor{darkred}{$+19.6$} & \textcolor{darkred}{$+20.9$} & \textcolor{red}{$+4.3$}  & \textcolor{red}{$+1.6$}  & \textcolor{red}{$+0.6$}  & \textcolor{red}{$+3.2$}  & \textcolor{red}{$+3.0$}  \\
\midrule
\multicolumn{11}{c}{\textbf{Image Classification (Accuracy $\uparrow$; $\Delta$ in pp)}} \\
\midrule
FMNIST & 200 & 500  & \textcolor{red}{$+1.0$}  & \textcolor{darkgreen}{$-0.2$}  & \textcolor{red}{$+2.5$}  & \textcolor{red}{$+2.7$}  & \textcolor{red}{$+4.6$}  & \textcolor{red}{$+4.6$}  & \textcolor{red}{$+4.6$}  & \textcolor{red}{$+3.4$}  \\
FMNIST & 200 & 2000 & \textcolor{red}{$+0.6$}  & \textcolor{darkgreen}{$-0.1$}  & \textcolor{red}{$+1.5$}  & \textcolor{red}{$+1.5$}  & \textcolor{red}{$+1.7$}  & \textcolor{red}{$+1.7$}  & \textcolor{red}{$+1.7$}  & \textcolor{red}{$+1.3$}  \\
CIFAR  & 200 & 500  & \textcolor{red}{$+2.2$}  & \textcolor{red}{$+0.2$}  & \textcolor{red}{$+2.3$}  & \textcolor{red}{$+2.3$}  & \textcolor{red}{$+3.1$}  & \textcolor{red}{$+3.1$}  & \textcolor{red}{$+3.1$}  & \textcolor{red}{$+2.3$}  \\
CIFAR  & 200 & 2000 & \textcolor{red}{$+2.3$}  & \textcolor{red}{$+0.6$}  & \textcolor{red}{$+2.2$}  & \textcolor{red}{$+2.4$}  & \textcolor{red}{$+2.7$}  & \textcolor{red}{$+2.7$}  & \textcolor{red}{$+2.7$}  & \textcolor{red}{$+2.2$}  \\
CelebA & 400 & 2000 & \textcolor{darkred}{$+6.4$}  & \textcolor{darkred}{$+10.0$} & \textcolor{darkred}{$+8.6$}  & \textcolor{darkred}{$+11.0$} & \textcolor{darkred}{$+11.0$} & \textcolor{darkred}{$+11.0$} & \textcolor{darkred}{$+18.0$} & \textcolor{darkred}{$+6.2$}  \\
CelebA & 400 & 4000 & \textcolor{darkred}{$+5.2$}  & \textcolor{darkred}{$+7.0$}  & \textcolor{darkred}{$+8.9$}  & \textcolor{darkred}{$+11.2$} & \textcolor{darkred}{$+11.2$} & \textcolor{darkred}{$+11.2$} & \textcolor{darkred}{$+13.4$} & \textcolor{darkred}{$+5.5$}  \\
\bottomrule
\end{tabular}}
\end{table*}

\subsection{Budget Scaling}
\label{app:budget_scaling}

\begin{figure}[t]
    \centering
    \begin{subfigure}{\linewidth}
        \centering
        \includegraphics[width=\linewidth]{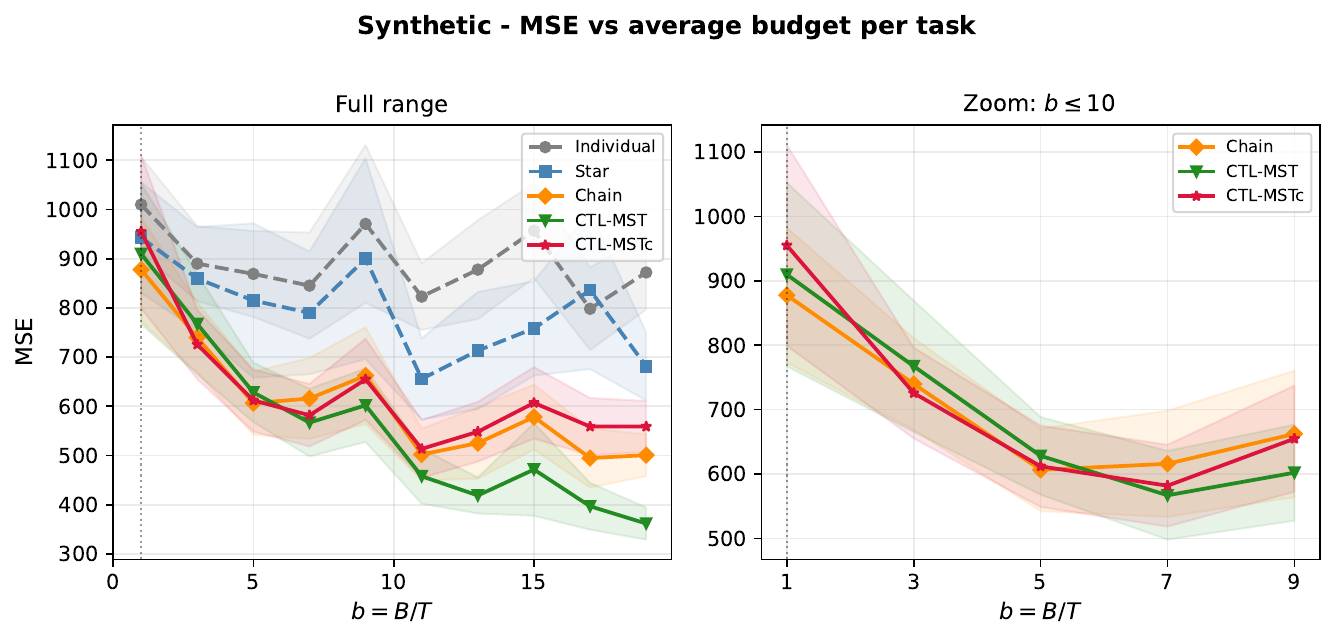}
        \caption{Synthetic regression (MSE).}
        \label{fig:bscale_syn}
    \end{subfigure}
    \begin{subfigure}{\linewidth}
        \centering
        \includegraphics[width=\linewidth]{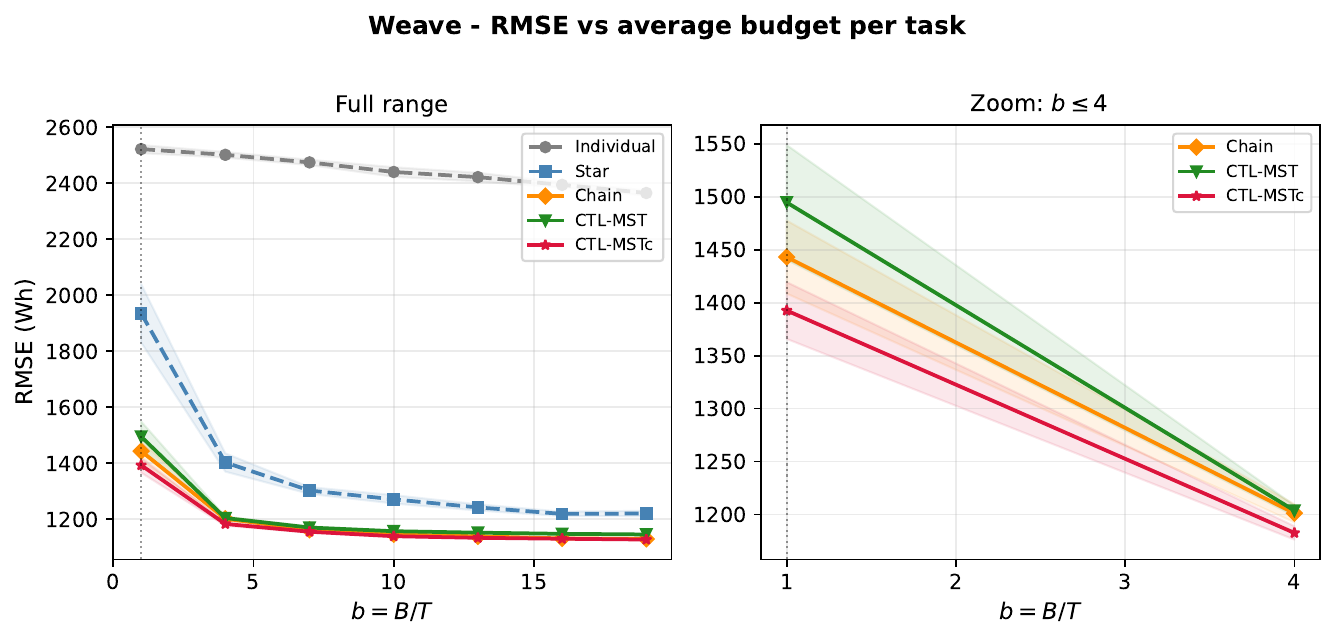}
        \caption{WEAVE-UK electricity (RMSE, Wh).}
        \label{fig:bscale_weave}
    \end{subfigure}
    \begin{subfigure}{\linewidth}
        \centering
        \includegraphics[width=\linewidth]{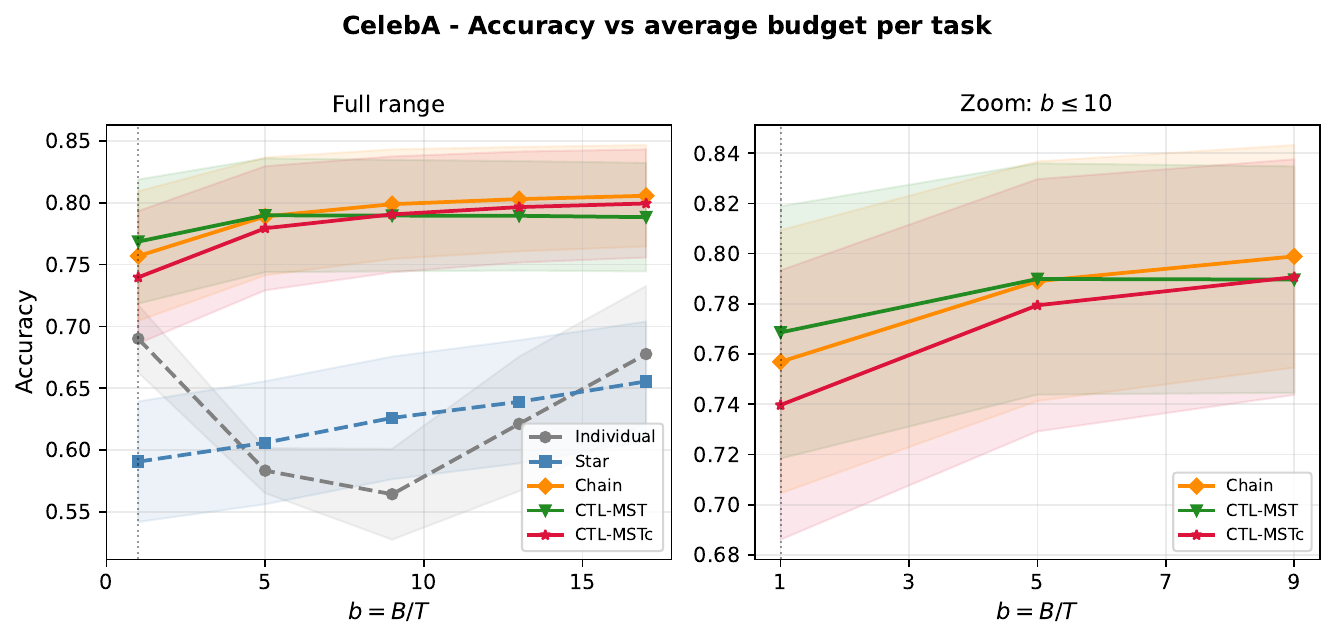}
        \caption{CelebA (accuracy, \%).}
        \label{fig:bscale_celeba}
    \end{subfigure}
    \caption{Test performance vs.\ average per-node budget $b = B/T$ (synthetic, WEAVE-UK, CelebA). Budget allocated via the SLE formula; shaded bands are 95\% confidence intervals over 20 seeds.}
    \label{fig:budget_scaling}
\end{figure}

Figure~\ref{fig:budget_scaling} reports test performance as a function of the average per-node budget $b = B/T$ for all datasets and methods, with budgets allocated via the SLE formula. The $x$-axis spans a wide range of $b$ values, from severely frugal to generous regimes. Shaded bands indicate 95\% confidence intervals over 20 seeds.

Across all datasets, CTL tree methods (CTL-MST and CTL-MSTc) improve monotonically with $b$ and maintain a consistent advantage over Star and Individual at every budget level. The gain is largest at low $b$, where warm-starting from a nearby task offsets the limited refinement capacity; it narrows but does not vanish as $b$ grows, indicating that the cascade structure provides a systematic rather than purely compensatory benefit. Chain remains competitive on near-one-dimensional geometries (FMNIST, CelebA) but falls behind tree methods on clustered datasets as $b$ increases, consistent with the depth-error analysis of Section~\ref{sec:construction}. Individual training saturates relatively early, reflecting the limited data available per task.

\begin{figure}[t]
    \centering
    \includegraphics[width=\linewidth]{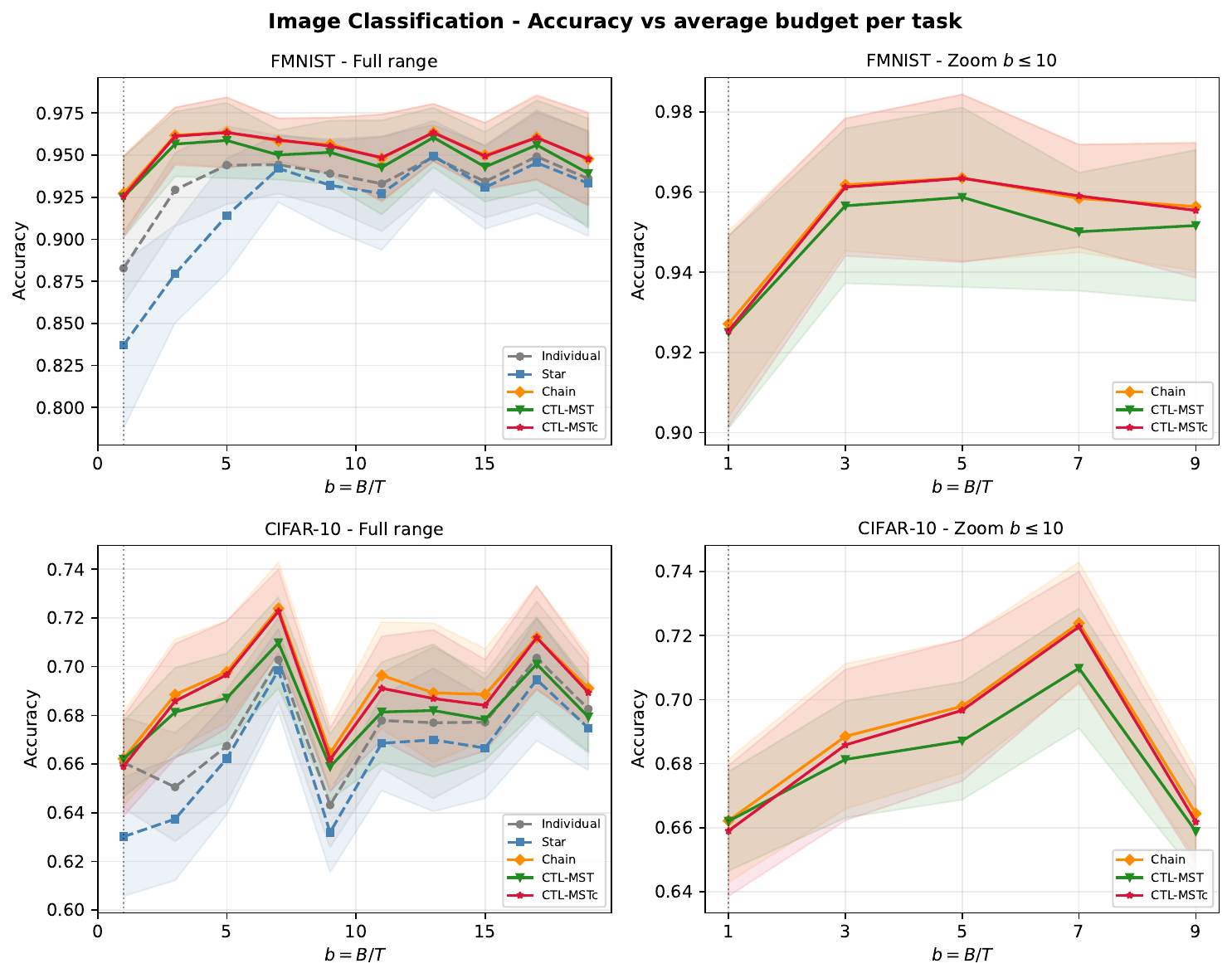}
    \caption{Test accuracy vs.\ average per-node budget $b = B/T$ for image classification (FMNIST and CIFAR-10). Budget allocated via the SLE formula; shaded bands are 95\% confidence intervals over 20 seeds.}
    \label{fig:bscale_classif}
\end{figure}

\subsection{Synthetic Regression}
\label{app:synthetic}

\begin{figure}[t]
    \centering
    \includegraphics[width=1.\linewidth]{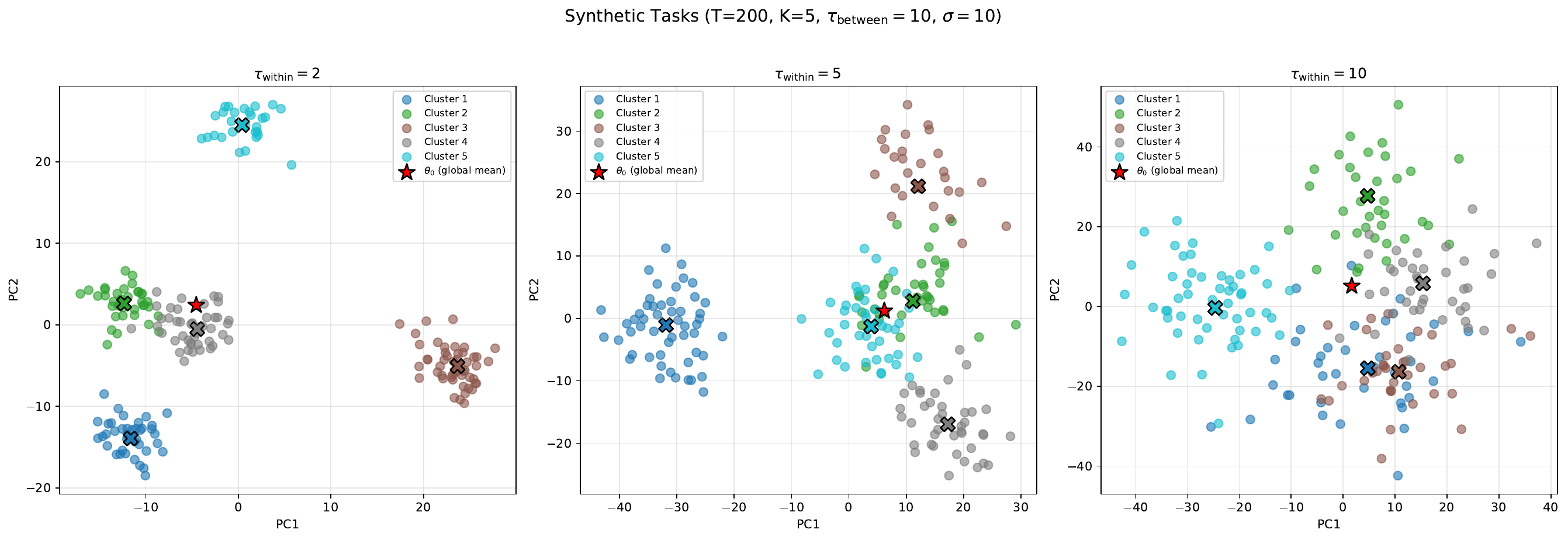}
    \caption{Two-dimensional PCA projection of task parameters for the synthetic data with increasing within-cluster variance $\tau_{\text{within}}$.}
    \label{fig:2d_clustered}
\end{figure}

Figure~\ref{fig:2d_clustered} illustrates the geometry of the synthetic tasks. Increasing the within-cluster variance $\tau_{\text{within}}$ progressively disperses task parameters while preserving the same between-cluster separation, resulting in a controlled degradation of task similarity.

\subsection{WEAVE-UK Electricity Forecasting}
\label{app:uk}

\begin{figure}[t]
    \centering
    \begin{subfigure}{0.48\linewidth}
        \centering
        \includegraphics[width=.7\linewidth]{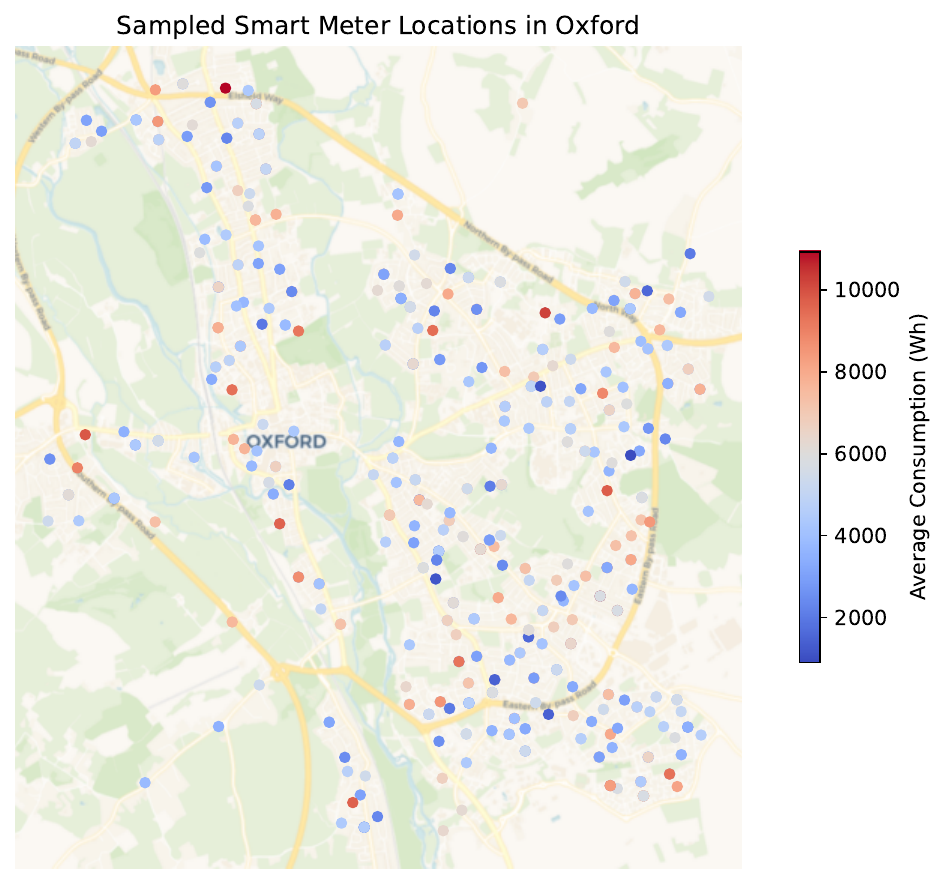}
        \caption{Spatial distribution of the sampled meters.}
        \label{fig:oxford_map}
    \end{subfigure}
    \hfill
    \begin{subfigure}{0.48\linewidth}
        \centering
        \includegraphics[width=.7\linewidth]{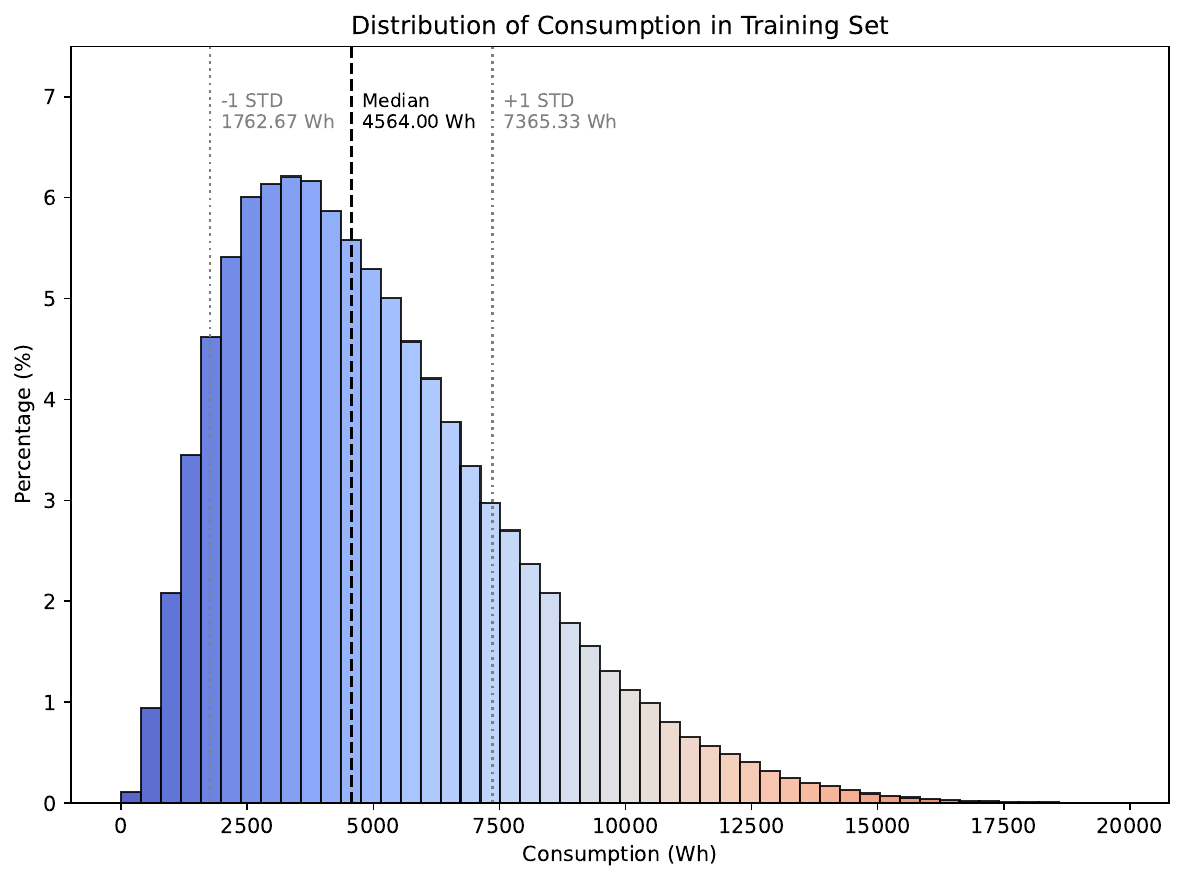}
        \caption{Distribution of yearly electricity consumption.}
        \label{fig:oxford_dist}
    \end{subfigure}
    \caption{Overview of spatial positions and consumption distribution of the sampled smart meters in Oxford's urban area.}
    \label{fig:oxford_extended}
\end{figure}

Figure~\ref{fig:oxford_extended} provides additional context on the WEAVE-UK electricity benchmark. The spatial distribution of the $500$ sampled smart meters shows broad geographic coverage, while the consumption histogram highlights a strongly skewed demand distribution with a heavy right tail.

\subsection{Image Classification Details}
\label{app:image_classif}

\paragraph{Feature extraction.}
For Fashion-MNIST and CIFAR-10, raw images are standardized (zero mean, unit variance per pixel) and then projected onto the top-$100$ principal components fitted on the full training set, yielding a $100$-dimensional input for each task. Each task is a binary classification problem over a pair of classes, with $n_{\mathrm{train}}=64$ and $n_{\mathrm{test}}=128$ samples per task. For CelebA, features are the $512$-dimensional penultimate activations of a pretrained ResNet-18 (ImageNet weights, frozen); no PCA reduction is applied.

\paragraph{Model and optimizer.}
All image classification tasks use a \emph{logistic regression} head (a single linear layer with sigmoid activation), optimized by gradient descent. Learning rates are selected per dataset via Optuna-based hyperparameter search (50 trials, 20 repetitions each): $\eta=0.296$ for Fashion-MNIST, $\eta=0.067$ for CIFAR-10, and $\eta$ is re-optimized for CelebA. $\ell_2$ regularization is applied with dataset-specific coefficients ($\lambda=7.9\times10^{-3}$ for FMNIST, $\lambda=4.4\times10^{-4}$ for CIFAR-10). The budget $B$ counts total gradient steps, distributed via the subtree-weighted log-edge formula $b_v = |s_v|\log\!\bigl(1+d(\mathrm{pa}(v),v)\bigr)$ (Appendix~\ref{app:budget_ablation}). All methods use identical architectures and hyperparameters.

\subsection{Pretrained Features and Foundation Models}
\label{app:pretrained}

We evaluate CTL's effectiveness when tasks are represented using powerful pretrained features, to assess whether the cascade structure adds value on top of strong initialization.

\paragraph{Frozen ResNet-18 (CIFAR-10 and Fashion-MNIST).}
We replace the raw PCA features with 512-d ImageNet embeddings from a frozen ResNet-18, and fine-tune a linear head per task. Task-specific PCA embeddings achieve higher absolute accuracy on Fashion-MNIST (CTL-MSTc: $93.3\%$ vs.\ $85.2\%$ at $B=500$), while ResNet is stronger on CIFAR-10 ($80.4\%$ vs.\ $63.8\%$). In both cases, CTL-MSTc improves over Star and Individual, as summarized in Table~\ref{tab:resnet}.

\begin{table}[t]
\centering
\tiny
\caption{Image classification accuracy (\%) with frozen ResNet-18 features (20 seeds, 95\% CI). $^*$ denotes significantly better than Star (one-sided Welch $t$-test, $p<0.05$).}
\label{tab:resnet}
\resizebox{\columnwidth}{!}{%
\begin{tabular}{llccc}
\toprule
Dataset & $B$ & Individual & Star & CTL-MSTc \\
\midrule
FMNIST   & 500  & $71.6\ [70.8, 72.4]$ & $74.6\ [73.6, 75.5]$ & $\mathbf{85.2\ [84.5, 85.9]}$$^*$ \\
FMNIST   & 2000 & $89.7\ [89.2, 90.3]$ & $88.8\ [88.2, 89.4]$ & $\mathbf{90.7\ [90.1, 91.2]}$$^*$ \\
CIFAR-10 & 500  & $73.9\ [73.3, 74.5]$ & $65.5\ [64.5, 66.4]$ & $\mathbf{80.4\ [80.0, 80.9]}$$^*$ \\
CIFAR-10 & 2000 & $83.7\ [83.4, 84.0]$ & $81.1\ [80.7, 81.5]$ & $\mathbf{84.7\ [84.4, 85.0]}$$^*$ \\
\bottomrule
\end{tabular}}
\end{table}

CTL-MSTc gains $+14.9$~pp over Star on CIFAR-10 at $B=500$ and $+10.6$~pp on FMNIST, with gains remaining significant at $B=2000$ ($+3.6$~pp and $+1.9$~pp). CTL-MST and CTL-MSTc both outperform RandTree in all settings, confirming that the improvement comes from geometry-aware cascade construction rather than tree structure alone.

\paragraph{Time-series foundation model (Chronos T5-small).}
We evaluate CTL on the WEAVE-UK electricity dataset using a frozen ChronosV2 T5-small backbone, fine-tuning per task either a linear probe or a LoRA adapter \citep{hu2022lora}, with rank $r=4$. As shown in Table~\ref{tab:chronos}, CTL-MSTc-gradient improves over Star and Individual in both regimes:

\begin{table}[t]
\centering
\caption{WEAVE-UK electricity RMSE with frozen Chronos T5-small backbone (linear probe and LoRA). Lower is better.}
\label{tab:chronos}
\begin{tabular}{llccc}
\toprule
Method & $B$ & Individual & Star & CTL-MSTc \\
\midrule
Linear probe & 1000 & $2243.2$ & $1748.3$ & $\mathbf{1625.8}$ \\
Linear probe & 2000 & $2183.3$ & $1658.6$ & $\mathbf{1595.3}$ \\
\midrule
LoRA ($r=4$) & 1000 & $2301.3$ & $1728.8$ & $\mathbf{1361.3}$ \\
LoRA ($r=4$) & 2000 & $2236.7$ & $1603.8$ & $\mathbf{1267.4}$ \\
\bottomrule
\end{tabular}
\end{table}

CTL consistently outperforms both Individual and Star in all configurations, confirming that cascade transfer organization generalizes across model classes. These results show that CTL can in principle be built on top of a foundation model. However, in our setting the Chronos linear probe (frozen features, linear head) falls short of the plain linear CTL baseline on raw features, despite consuming substantially more compute ($>8$ min per LoRA run vs.\ ${\sim}30$ s for the linear baseline). This suggests that, at least for this dataset and budget regime, the FM backbone does not provide a representation advantage over raw lag features, and the compute overhead is not justified. CTL therefore delivers its strongest gains in the linear regime, where transfer bias is the binding constraint rather than representation quality.

\section{Budget Allocation and Construction Hyperparameters}
\label{app:budget_ablation}

\paragraph{Choice of $\phi_\tau$ and $\psi$.}
In all experiments we use $\phi_{\tau}(d)=\log\!\big(1+\exp((d-\tau)/\tau)\big)$ (softplus centered at the median pairwise distance $\tau$), which provides a smooth transition between the near-field ($d\ll\tau$, low penalty) and far-field ($d\gg\tau$, linear penalty) regimes. Alternative choices such as the hinge $\phi_\tau(d)=\max(0,d-\tau)$ or the identity $\phi_\tau(d)=d$ are suitable when the task distance distribution is more uniform or heavy-tailed. 
For the subtree penalty we use $\psi(m)=m$ (linear in subtree size); alternatives such as $\psi(m)=\log(1+m)$ (softer discounting of large subtrees) or $\psi(u)=\deg_{\mathcal{T}}(u)+1$ (degree-based, discouraging high branching factor at individual nodes) can be better suited to geometries where greedy attachment tends to create unbalanced trees with deep single-branch paths.

\paragraph{Budget allocation.}
The global budget $B$ is distributed across tasks via $b_v = |s_v|\log\!\bigl(1+d(\mathrm{pa}(v),v)\bigr)$, where $|s_v|$ is the subtree size rooted at $v$ and $d(\mathrm{pa}(v),v)$ is the edge length to its parent. This formula jointly accounts for structural position (nodes responsible for larger subtrees receive more budget) and transfer difficulty (longer edges, which carry higher transfer bias, attract additional refinement effort). We compared this formula against a pure uniform allocation in a controlled synthetic sweep over $\tau_{\text{within}} \in \{2,5,10\}$ and $B \in \{500,1000,2000\}$ (50 random draws each). The subtree-weighted log-edge formula consistently matched or improved upon uniform allocation, with gains most pronounced at low budget and low-to-moderate heterogeneity, where the structure of the tree provides reliable signal for redistribution. At high heterogeneity ($\tau_{\text{within}}=10$), differences between the two schemes narrow, and neither compensates for weak task relatedness. The selected formula is therefore used as the default in all experiments.

\paragraph{Theoretical budget derivation.}
\label{app:budget_derivation}
We derive the optimal budget allocation by minimizing a tractable surrogate of the total cascade error under a mean-field approximation.
From Proposition~\ref{prop:propagation-param}, the error at task $v$ depends on all ancestor edges. Summing over all tasks $v \in \mathcal{V}$ and exchanging the order of summation, an edge $u$ contributes to the error of every descendant $v \in s_u$:
\begin{equation*}
\mathcal{E}\bigl(\{b_v\}\bigr) = \sum_{v\in\mathcal V} \left\lVert\tilde\bT_v^{\mathrm{_\textsc{CTL}}}-\bT_v^\star\right\rVert
= \sum_{u \in \mathcal{V}} d\bigl(\mathrm{pa}(u), u\bigr)
  \sum_{v \in s_u}
  \prod_{w \in \mathrm{path}(u \to v)} \rho_w^{b_w}.
\end{equation*}

The inner sum couples all nodes along each path to $v$. To obtain a closed-form allocation, we make two simplifications: (i) we assume a uniform contraction rate $\rho_v = \rho \in (0,1)$ for all $v \in \mathcal{V}$, and (ii) we retain only the leading contraction at node $u$, replacing the full path product by $\rho^{b_u}$:
\begin{equation*}
\sum_{v \in s_u} \prod_{w \in \mathrm{path}(u \to v)} \rho^{b_w} \;\approx\; |s_u|\,\rho^{b_u},
\end{equation*}
where $|s_u|$ is the subtree size of $u$. This mean-field approximation is accurate when the tree is shallow below $u$ (as in a balanced tree of depth $O(\log T)$) or when downstream budgets are comparable to $b_u$; it becomes less accurate in deep chains where the path product $\prod_w \rho^{b_w}$ compounds over many nodes. This yields the separable surrogate
\begin{equation*}
\tilde{\mathcal{E}}\bigl(\{b_v\}\bigr) = \sum_{u \in \mathcal{V}} w_u\, e^{-\kappa b_u}, \qquad w_u = |s_u|\,d\bigl(\mathrm{pa}(u),u\bigr),\quad \kappa = -\ln\rho > 0.
\end{equation*}

Minimizing $\tilde{\mathcal{E}}$ subject to $\sum_u b_u = B$ and $b_u \geq 0$, we apply the KKT conditions. At an interior solution ($b_u > 0$), complementary slackness forces the non-negativity multiplier $\nu_u = 0$, and the stationarity condition gives
\begin{equation*}
-\kappa w_u e^{-\kappa b_u^\star} + \lambda = 0
\;\implies\;
b_u^\star = \frac{1}{\kappa}\ln\!\left(\frac{\kappa w_u}{\lambda}\right).
\end{equation*}
Enforcing $\sum_u b_u^\star = B$ determines $\ln\lambda = \tfrac{1}{|\mathcal{V}|}\sum_{u'}\ln(\kappa w_{u'}) - \tfrac{\kappa B}{|\mathcal{V}|}$. Substituting back:

\begin{equation}
b_u^\star = \frac{B}{|\mathcal{V}|} + \frac{1}{\kappa}\!\left(\ln \Big(|s_u|\,d\bigl(\mathrm{pa}(u),u\bigr)\Big) - \frac{1}{|\mathcal{V}|}\sum_{u' \in \mathcal{V}} \ln\Big(|s_{u'}|\,d\bigl(\mathrm{pa}(u'),u'\bigr)\Big)\right).
\label{eq:water-filling}
\end{equation}

Each node receives the uniform budget $B/|\mathcal{V}|$ corrected by how its log-importance $\ln w_u$ deviates from the mean. Nodes with large subtrees or long parent edges receive more budget; nodes deep in a chain with short edges receive less. If the formula yields $b_u^\star < 0$ for some nodes, those are clamped to a minimum floor and the residual budget is redistributed proportionally among the remaining nodes.

\paragraph{Connection to the SLE formula.}
Equation~\eqref{eq:water-filling} minimizes the mean-field surrogate $\tilde{\mathcal{E}}$ and yields an additive decomposition in log-space: $b_v^\star \propto \log|s_v| + \log d(\mathrm{pa}(v),v)$.
The SLE formula $b_v \propto |s_v|\log(1+d(\mathrm{pa}(v),v))$ used in all experiments adopts a different principle: it equalizes the post-refinement residual $\rho^{b_v} d(\mathrm{pa}(v),v)$ across edges and scales by subtree size $|s_v|$ to allocate proportionally more budget to edges with high downstream leverage. This multiplicative weighting gives stronger emphasis to large-subtree nodes than the additive KKT formula, and performs better for regression tasks where the mean-field cascade error is quadratic. For classification tasks, the additive KKT formula performs better; see the ablation in Table~\ref{tab:alloc_ablation}.

\paragraph{Structure vs.\ allocation.}
To separate the contribution of tree structure from budget allocation we fix the tree to CTL-MST-gradient and vary the allocation across three schemes: uniform ($b_v = (B-b_s)/(T-1)$ for all non-root nodes), KKT-additive ($b_v \propto \log|s_v|+\log(1+d(\mathrm{pa}(v),v))$), and SLE-multiplicative ($b_v \propto |s_v|\log(1+d(\mathrm{pa}(v),v))$). Table~\ref{tab:alloc_ablation} reports means over 20 seeds.

\begin{table}[t]
\centering
\caption{%
  Attribution of gains into tree-structure and allocation effects (means over 20 seeds).
  $\Delta_{\text{struct}} = \text{MST-uniform} - \text{Star}$ isolates the topology benefit at fixed uniform allocation.
  $\Delta_{\text{KKT}}$ and $\Delta_{\text{SLE}}$ are the further increments from switching allocation while holding the MST fixed.
  Negative = improvement for RMSE$\downarrow$; positive = improvement for Acc$\uparrow$.}
\label{tab:alloc_ablation}
\small
\begin{tabular}{llrrr}
\toprule
Dataset & $B$ & $\Delta_{\text{struct}}$ & $\Delta_{\text{KKT}}$ & $\Delta_{\text{SLE}}$ \\
\midrule
\multirow{3}{*}{Syn-10 (RMSE$\downarrow$)}
 & 500  & $+0.22$ & $+0.42$ & $-0.25$ \\
 & 1000 & $-0.36$ & $-1.40$ & $-2.37$ \\
 & 2000 & $-1.03$ & $-5.93$ & $-6.66$ \\
\midrule
\multirow{3}{*}{WEAVE-UK (RMSE$\downarrow$)}
 & 500  & $-188.3$ & $-56.1$ & $-326.3$ \\
 & 1000 & $-106.4$ & $-89.2$ & $-352.8$ \\
 & 2000 & $-141.5$ & $-70.2$ & $-199.0$ \\
\midrule
\multirow{3}{*}{FMNIST (Acc$\uparrow$)}
 & 500  & $+4.2$ & $+17.2$ & $+13.8$ \\
 & 1000 & $+5.4$ & $+16.4$ & $+13.4$ \\
 & 2000 & $+9.3$ & $+12.8$ & $+10.4$ \\
\midrule
\multirow{3}{*}{CIFAR-10 (Acc$\uparrow$)}
 & 500  & $+4.0$ & $+2.9$ & $-0.5$ \\
 & 1000 & $+4.1$ & $+3.0$ & $-0.4$ \\
 & 2000 & $+4.2$ & $+3.0$ & $-0.3$ \\
\bottomrule
\end{tabular}
\end{table}

Three findings emerge. \textbf{(i)}~\emph{Tree structure is consistently beneficial}: replacing the star topology with CTL-MST-uniform improves performance on all benchmarks at $B \ge 1000$ and substantially at all budgets on WEAVE-UK ($\Delta_{\text{struct}} \ge 106$ RMSE points) and CIFAR-10/FMNIST ($+4$--$9$ accuracy points). At $B=500$ on synthetic tasks the effect reverses slightly ($+0.22$ RMSE), because the per-node budget $b\approx 2$ is too small to refine away the cascade bias introduced by MST edges. \textbf{(ii)}~\emph{SLE allocation dominates for regression}: holding the MST fixed, switching from uniform to SLE provides an additional reduction of 0.25--6.66 RMSE on synthetic data and 199--353 RMSE on WEAVE-UK, exceeding the tree-structure benefit at $B \ge 1000$. The allocation effect grows with $B$ because larger budgets allow the exponential decay $\rho^{b_v}$ in the error surrogate to be steered by allocation more decisively. \textbf{(iii)}~\emph{KKT allocation is consistently best for classification}: KKT gains $+12.8$--$17.2$ pp on FMNIST and $+2.9$--$3.0$ pp on CIFAR-10 over uniform. SLE also helps on FMNIST ($+10.4$--$13.8$ pp, less than KKT), but hurts on CIFAR-10 ($-0.3$ to $-0.5$ pp).
\section{Additional Diagnostic Results}
\label{app:diagnostics}

This section reports compact diagnostics supporting claims in the main text on root sensitivity, proxy quality, timing overhead, and effective budget accounting.

\paragraph{Graph construction overhead.}
\begin{table}[t]
\centering
\caption{Graph construction time vs.\ training time for the gradient distance with $B=2000$ over $50$ seeds.}
\label{tab:timing}
\small
\begin{tabular}{ccccc}
\toprule
$T$ & Graph time & Training time & Ratio \\
\midrule
$100$ & $0.001 \pm 0.000$\,s & $1.107 \pm 0.011$\,s & $0.09\%$ \\
$200$ & $0.004 \pm 0.000$\,s & $1.096 \pm 0.006$\,s & $0.37\%$ \\
$1000$ & $0.126 \pm 0.002$\,s & $0.827 \pm 0.008$\,s & $15.2\%$ \\
\bottomrule
\end{tabular}
\end{table}

\paragraph{Root sensitivity.} On synthetic regression tasks, we compare three rooting strategies for the MST: medoid (our default), random root, and depth-central (the node minimizing the maximum path length). Mean MSE across configurations is $587.3 \pm 51.4$ for depth-central, $618.0 \pm 61.5$ for medoid, and $629.2 \pm 60.5$ for random rooting. Depth-central rooting produces shallower, more balanced trees that reduce error accumulation, while the medoid offers a good data-driven compromise without requiring explicit tree traversal optimization.

\paragraph{Proxy quality.} We audit proxy quality by measuring average gain over Star on synthetic regression tasks. The gradient proxy yields $+31.7\%$ average gain, while the feature and target proxies achieve more modest gains ($+18.1\%$ and $+17.1\%$ respectively), consistent with the main experimental findings. Note that element-wise feature and target distances are not well-defined on tasks with independently sampled instances (as in our synthetic and image benchmarks), since row ordering is arbitrary across tasks; their gains should be interpreted as lower bounds on what an aligned variant might achieve.

\paragraph{Depth--error correlation does not grow with scale.}
If noise accumulated strongly through the cascade, the Spearman correlation $\rho(\text{depth}, Q)$ between a task's depth and its test quality $Q$ (MSE for Synth, Accuracy for CelebA) should increase with $T$, since more tasks lead to deeper trees.
Table~\ref{tab:depth_corr_vs_T} shows this correlation across scales, on both the synthetic regression setting ($\tau_{\text{within}}=10$, $B=5T$) and on CelebA binary classification ($B=5T$, gradient proxy), with 10 runs per configuration.

\begin{table}[t]
\centering
\caption{Spearman $\rho(\text{depth}, Q)$ across task scales. $Q=\text{MSE}$ for Synth (lower = better, so positive $\rho$ means deeper nodes are \emph{worse}); $Q=\text{Accuracy}$ for CelebA (higher = better, so negative $\rho$ means deeper nodes are \emph{worse}). Values near zero indicate no systematic depth-driven error accumulation.}
\label{tab:depth_corr_vs_T}
\small
\begin{tabular}{ccccc}
\toprule
Dataset & $T$ & $\rho$ & Max depth & Mean depth \\
\midrule
\multirow{7}{*}{Synth} & 300  & $-0.198 \pm 0.129$ & $41.2 \pm 7.6$   & $21.2 \pm 3.9$ \\
      & 500  & $-0.092 \pm 0.102$ & $57.4 \pm 8.3$   & $31.3 \pm 6.2$ \\
      & 1000 & $-0.121 \pm 0.120$ & $75.8 \pm 9.7$   & $42.1 \pm 5.7$ \\
      & 2000 & $+0.019 \pm 0.062$ & $85.0 \pm 15.2$  & $45.4 \pm 9.7$ \\
      & 3000 & $+0.014 \pm 0.049$ & $123.8 \pm 22.0$ & $64.5 \pm 14.1$ \\
      & 4000 & $+0.043 \pm 0.046$ & $125.5 \pm 25.3$ & $70.7 \pm 18.3$ \\
      & 5000 & $+0.057 \pm 0.033$ & $136.1 \pm 17.9$ & $72.3 \pm 13.8$ \\
\midrule
\multirow{5}{*}{CelebA} & 100  & $+0.110 \pm 0.458$ & $17.8 \pm 5.5$  & $9.2 \pm 3.2$ \\
       & 200  & $+0.061 \pm 0.437$ & $23.8 \pm 5.8$  & $12.2 \pm 3.5$ \\
       & 300  & $-0.054 \pm 0.376$ & $25.4 \pm 3.6$  & $13.2 \pm 1.7$ \\
       & 400  & $+0.077 \pm 0.376$ & $31.3 \pm 10.5$ & $16.2 \pm 6.2$ \\
       & 1000 & $-0.024 \pm 0.327$ & $37.5 \pm 10.8$ & $19.7 \pm 5.5$ \\
\bottomrule
\end{tabular}
\end{table}

\begin{figure}[t]
    \centering
    \begin{subfigure}[b]{0.48\linewidth}
        \centering
        \includegraphics[width=\linewidth]{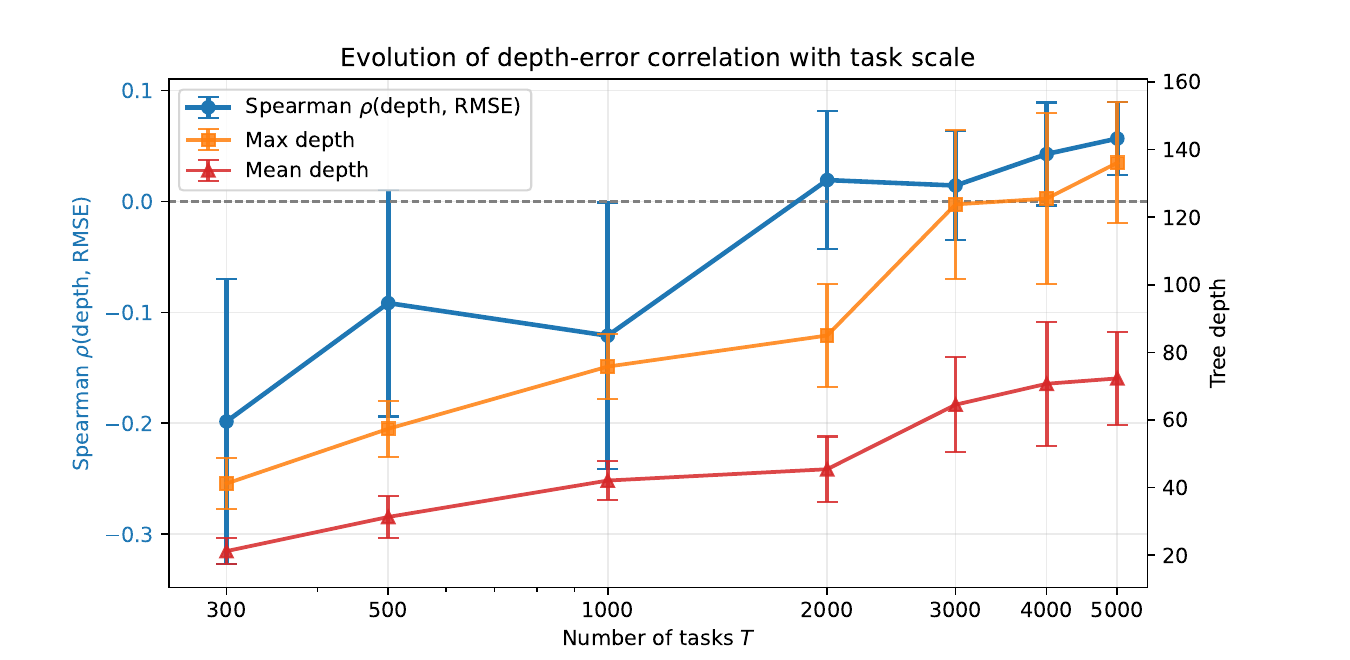}
        \caption{Synthetic regression. Blue: $\rho(\text{depth},\text{MSE})$. Orange: max depth. Despite depth growing to 136, correlation stays near zero.}
        \label{fig:path_corr_synth}
    \end{subfigure}
    \hfill
    \begin{subfigure}[b]{0.48\linewidth}
        \centering
        \includegraphics[width=\linewidth]{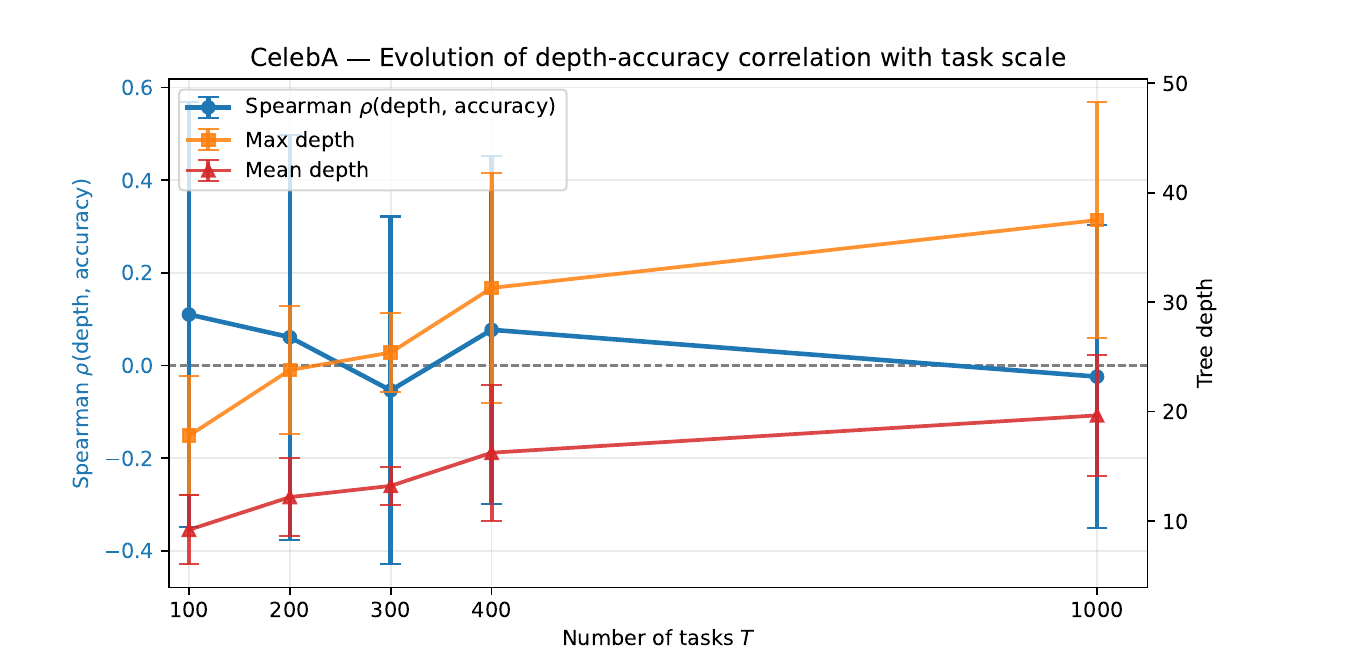}
        \caption{CelebA classification. Blue: $\rho(\text{depth},\text{Acc})$. Orange: max depth. No upward trend in correlation despite growing depth.}
        \label{fig:path_corr_celeba}
    \end{subfigure}
    \caption{Evolution of depth--quality Spearman correlation with task scale ($B=5T$, gradient proxy, 10 runs). The correlation stays near zero on both datasets as $T$ grows, confirming no practical depth-driven error accumulation.}
    \label{fig:path_corr_evolution}
\end{figure}

Despite maximum tree depth growing from 41 to 136 on Synth, and from 18 to 38 on CelebA, $\rho$ remains near zero throughout and shows no upward trend. Figure~\ref{fig:path_corr_evolution} illustrates this jointly with tree depth for both datasets. These results indicate that noise does not compound with scale in practice.

\paragraph{CTL outperforms independent training at every depth level.}
To test whether deep nodes specifically suffer from noise accumulation, we compare per-depth-bucket performance of CTL versus independent training (IT, same budget $B/T$ per task, no transfer).
Table~\ref{tab:depth_bucket_gain} reports the gain (positive = CTL wins) pooled over multiple values of $T$.

\begin{table}[t]
\centering
\caption{CTL gain over independent training by depth bucket. Synth: gain = MSE$_\text{IT}$ $-$ MSE$_\text{CTL}$ (pooled over $T\in\{500,1000,2000,3000\}$, 65k nodes). CelebA: gain = Acc$_\text{CTL}$ $-$ Acc$_\text{IT}$ (pooled over $T\in\{100,200,300,400,1000\}$, 19k nodes). Mean $\pm$ SE over 10 runs. Positive values indicate CTL superiority.}
\label{tab:depth_bucket_gain}
\small
\begin{tabular}{cccr}
\toprule
Dataset & Depth bucket & \# nodes & CTL gain \\
\midrule
\multirow{5}{*}{Synth (MSE $\downarrow$)}
  & 0 (root) & 40     & $+21.88 \pm 2.75$ \\
  & 1        & 88     & $+15.58 \pm 2.87$ \\
  & 2        & 106    & $+14.37 \pm 2.85$ \\
  & 3--5     & 431    & $+12.80 \pm 2.68$ \\
  & $\ge 6$  & 64,335 & $+9.23 \pm 2.80$  \\
\midrule
\multirow{5}{*}{CelebA (Acc $\uparrow$)}
  & 0 (root) & 50     & $+0.243 \pm 0.031$ \\
  & 1        & 142    & $+0.211 \pm 0.035$ \\
  & 2        & 296    & $+0.214 \pm 0.037$ \\
  & 3--5     & 1,666  & $+0.207 \pm 0.041$ \\
  & $\ge 6$  & 16,846 & $+0.179 \pm 0.042$ \\
\bottomrule
\end{tabular}
\end{table}

CTL outperforms independent training at every depth level on both datasets. The gain decreases slightly with depth (consistent with mild error propagation as predicted by Proposition~\ref{prop:noisy_propagation}), but never becomes negative. Nodes at depth $\ge 6$, which constitute approximately 99\% of all nodes at large $T$, still benefit significantly: $+9.2$ MSE units on Synth and $+17.9$ pp accuracy on CelebA.

\begin{figure}[t]
    \centering
    \begin{subfigure}[b]{0.45\linewidth}
        \centering
        \includegraphics[width=\linewidth]{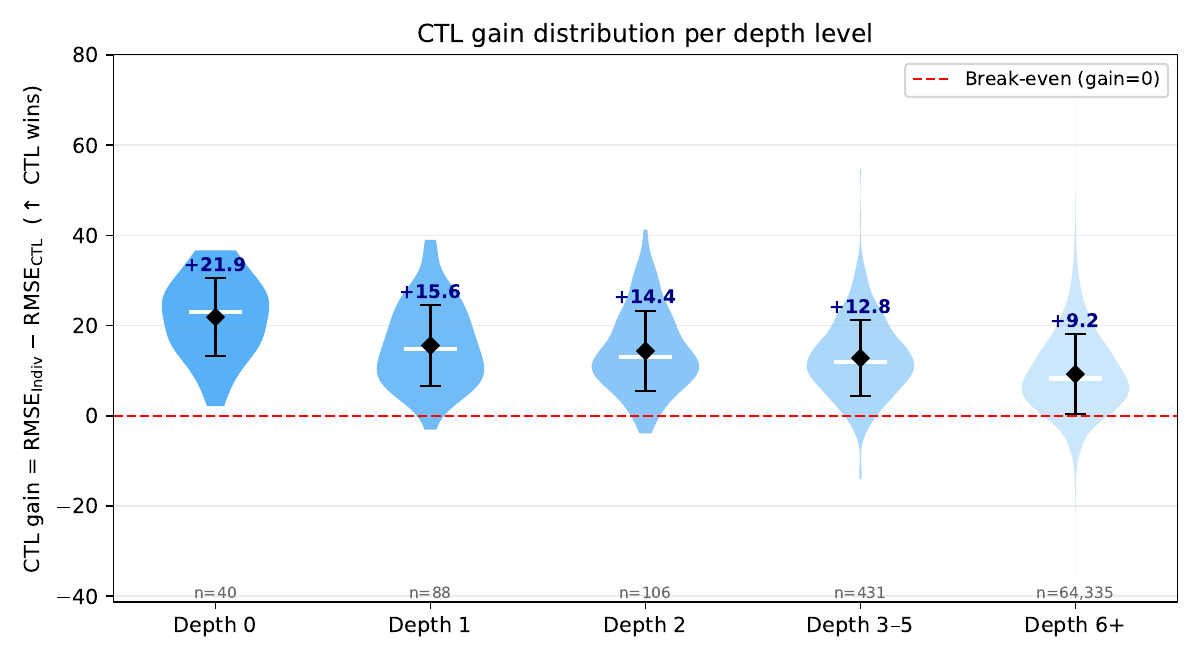}
        \caption{Synthetic: CTL gain (MSE$_\text{IT} -$ MSE$_\text{CTL}$) by depth bucket. Gains remain positive at all depths.}
        \label{fig:depth_gain_violin_synth}
    \end{subfigure}
    \hfill
    \begin{subfigure}[b]{0.45\linewidth}
        \centering
        \includegraphics[width=\linewidth]{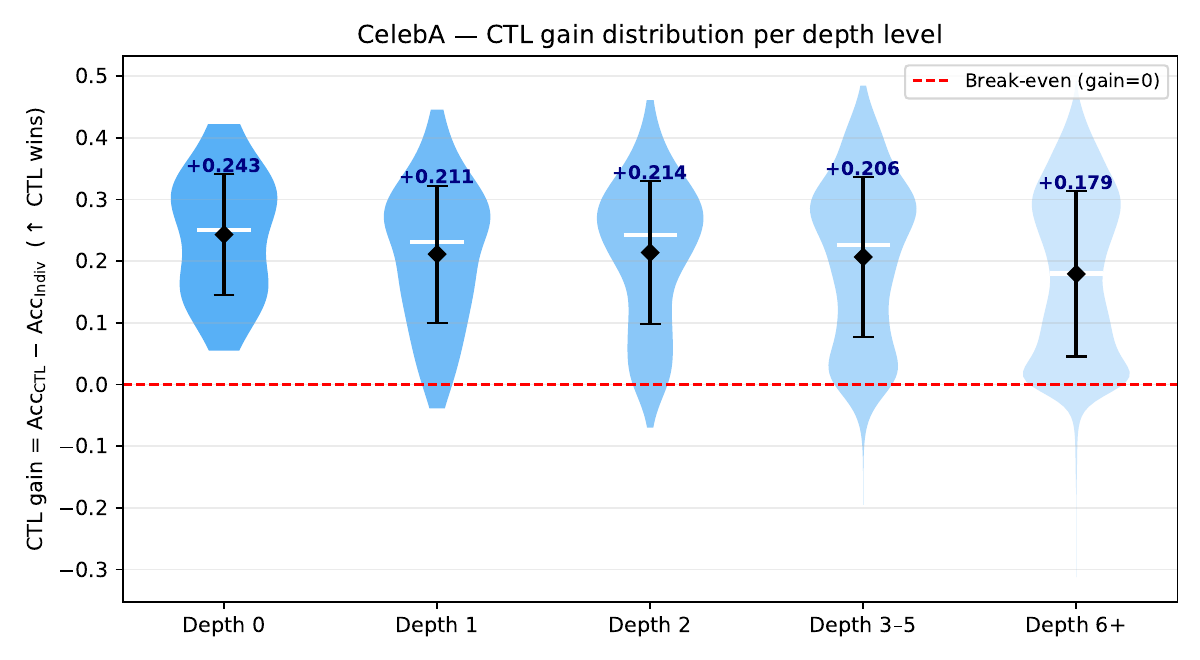}
        \caption{CelebA: CTL gain (Acc$_\text{CTL} -$ Acc$_\text{IT}$) by depth bucket. Same pattern: gains are always positive.}
        \label{fig:depth_gain_violin_celeba}
    \end{subfigure}
    \caption{Distribution of CTL gain over independent training by depth bucket (pooled over multiple values of $T$). Gains are positive at every depth level on both datasets.}
    \label{fig:depth_gain_violin}
\end{figure}

\begin{figure}[t]
    \centering
    \begin{subfigure}[b]{0.45\linewidth}
        \centering
        \includegraphics[width=\linewidth]{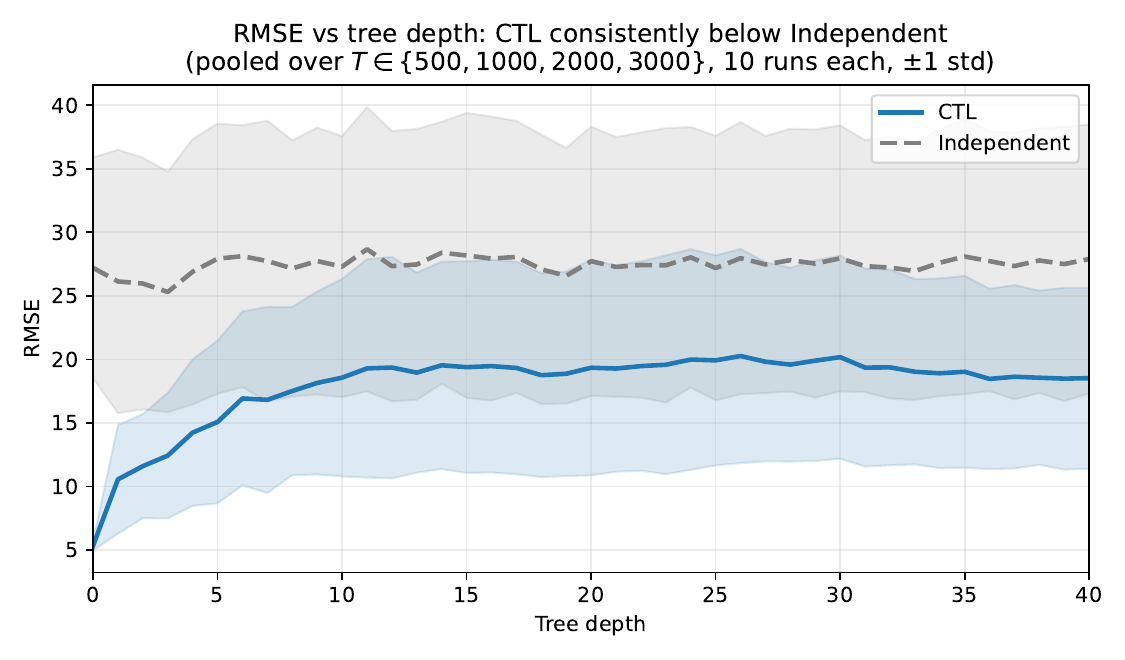}
        \caption{Synthetic: MSE vs.\ depth for CTL and IT across values of $T$. CTL is consistently below IT.}
        \label{fig:rmse_vs_depth_synth}
    \end{subfigure}
    \hfill
    \begin{subfigure}[b]{0.45\linewidth}
        \centering
        \includegraphics[width=\linewidth]{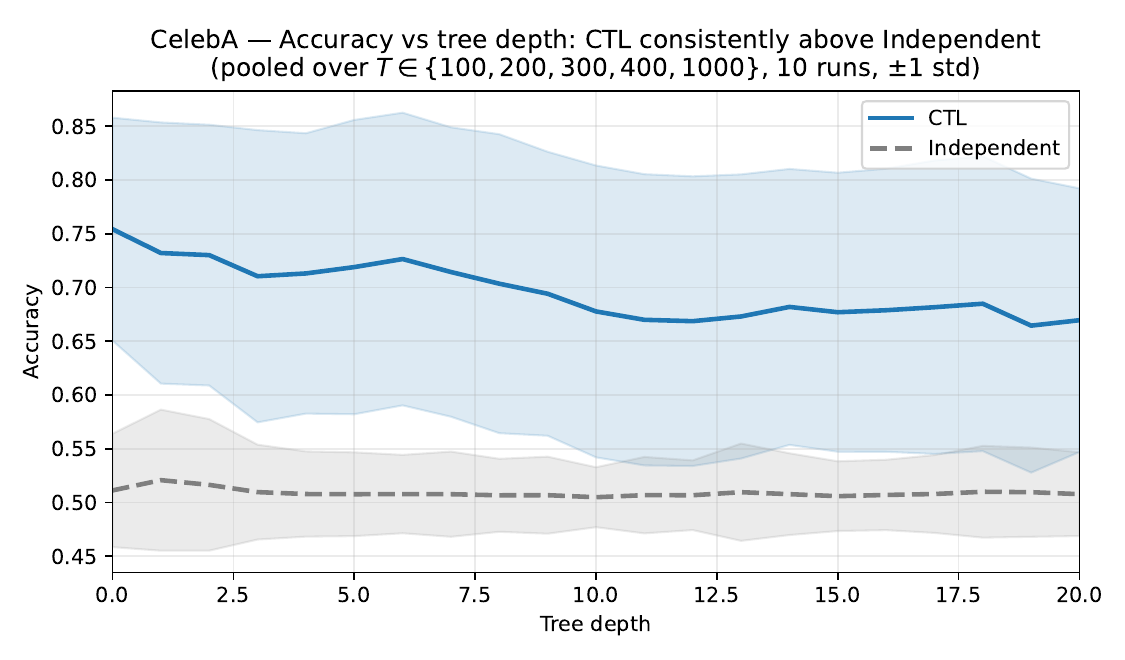}
        \caption{CelebA: Accuracy vs.\ depth for CTL and IT across values of $T$. CTL is consistently above IT.}
        \label{fig:acc_vs_depth_celeba}
    \end{subfigure}
    \caption{Test quality as a function of cascade depth, comparing CTL and independent training (IT). CTL retains its advantage at all depth levels on both datasets.}
    \label{fig:depth_quality_vs_depth}
\end{figure}

\begin{figure}[t]
    \centering
    \begin{subfigure}[b]{\linewidth}
        \centering
        \includegraphics[width=\linewidth]{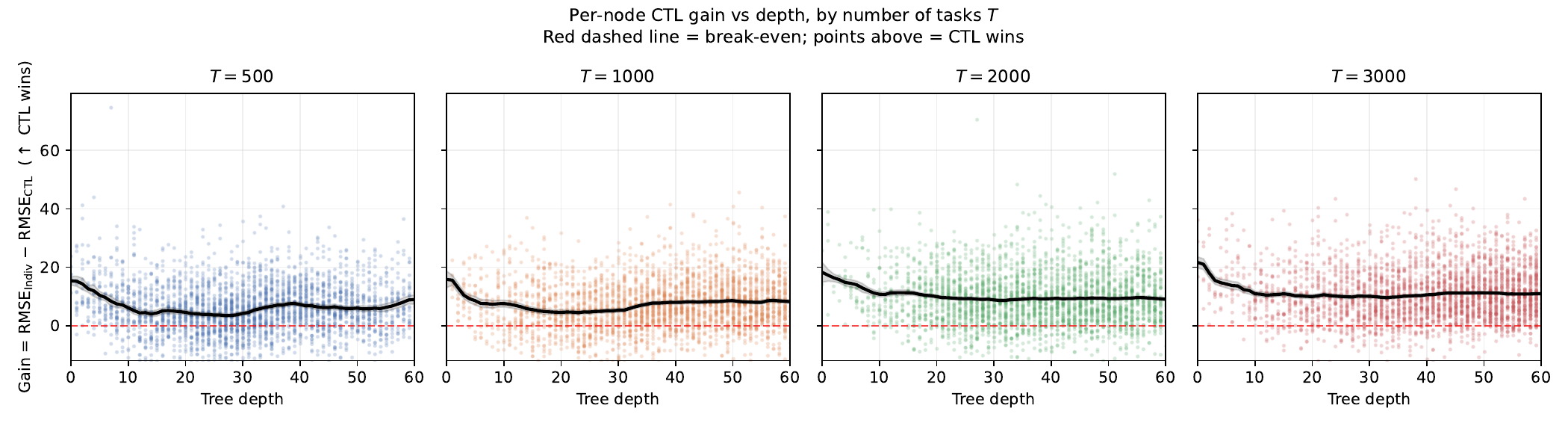}
        \caption{Synthetic: scatter of per-task CTL gain vs.\ depth, colored by $T$. Points are above zero for all depths and all $T$.}
        \label{fig:depth_scatter_synth}
    \end{subfigure}
    \\
    \begin{subfigure}[b]{\linewidth}
        \centering
        \includegraphics[width=\linewidth]{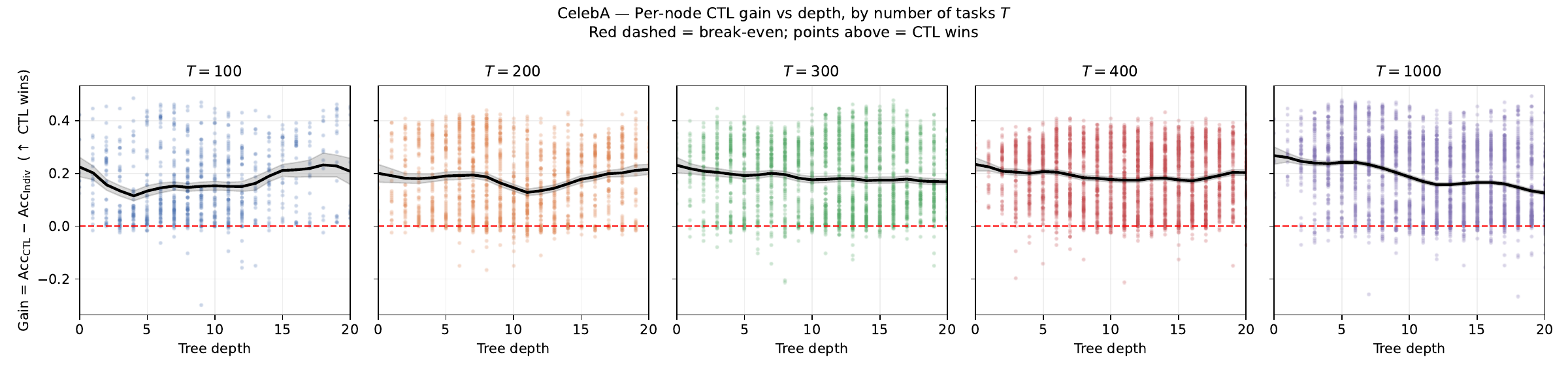}
        \caption{CelebA: same scatter for accuracy gain. No depth threshold beyond which CTL loses.}
        \label{fig:depth_scatter_celeba}
    \end{subfigure}
    \caption{Per-task CTL gain vs.\ cascade depth, colored by number of tasks $T$. CTL dominates independent training uniformly across depths and scales.}
    \label{fig:depth_scatter}
\end{figure}

Figures~\ref{fig:depth_gain_violin} and~\ref{fig:depth_quality_vs_depth} provide visual confirmation. The violin plots show the distribution of CTL gains at each depth bucket; gains remain positive and well above zero across all buckets. The quality-vs-depth curves show that CTL consistently tracks below (Synth) or above (CelebA) independent training at every depth level and for every tested value of $T$. The scatter plots (Figure~\ref{fig:depth_scatter}) confirm that this holds uniformly across task scales.

\end{document}